%% file: main_arixiv.tex
\documentclass[10pt,letterpaper]{article}

\usepackage[T1]{fontenc}
\usepackage[utf8]{inputenc}
\usepackage{times}

\usepackage[
  letterpaper,
  textwidth=5.5in,
  textheight=9in,
  centering,
  footskip=26pt
]{geometry}

\usepackage{titlesec}
\titleformat{\section}
  {\normalfont\large\scshape}
  {\thesection}{1em}{}
\titleformat{\subsection}
  {\normalfont\normalsize\scshape}
  {\thesubsection}{1em}{}
\titleformat{\subsubsection}
  {\normalfont\normalsize\scshape}
  {\thesubsubsection}{1em}{}

\titlespacing*{\section}
  {0pt}{2ex plus .5ex minus .2ex}{1.5ex plus .3ex}
\titlespacing*{\subsection}
  {0pt}{1.8ex plus .5ex minus .2ex}{.8ex plus .2ex}
\titlespacing*{\subsubsection}
  {0pt}{1.5ex plus .5ex minus .2ex}{.5ex plus .2ex}

\usepackage[authoryear,round]{natbib}
\setcitestyle{
  authoryear,
  round,
  citesep={;},
  aysep={,},
  yysep={;}
}

\date{}

\input{math_commands.tex}

\usepackage{float}
\usepackage{threeparttable}
\usepackage{url}
\usepackage{booktabs}
\usepackage{longtable}
\usepackage{amsthm}
\usepackage[table]{xcolor}
\usepackage{multirow}
\usepackage{makecell}
\usepackage{amsfonts}
\usepackage{nicefrac}
\usepackage{microtype}
\usepackage{graphicx}
\usepackage{amsmath,amssymb}
\usepackage{tabularx}
\usepackage{CJKutf8}

\usepackage{hyperref}
\hypersetup{
  hidelinks,
  pdftitle={HIFI: A Highly Efficient Framework for Statistical Inference with Foundation Models},
  pdfauthor={Po-Wei Chen, Shan Yu, Chunlin Li}
}

\definecolor{hifishade}{rgb}{0.937,0.957,0.969}

\newtheorem{theorem}{Theorem}[section]
\newtheorem{proposition}[theorem]{Proposition}
\newtheorem{lemma}[theorem]{Lemma}
\newtheorem{assumption}[theorem]{Assumption}

\title{HIFI: A Highly Efficient Framework for\\
Statistical Inference with Foundation Models}

\author{%
  Po-Wei Chen \quad Shan Yu \quad
  Chunlin Li\thanks{Corresponding author:
  \texttt{gjj2pr@virginia.edu}.}\\[6pt]
  {\small Department of Statistics, University of Virginia}\\
  {\small Charlottesville, VA, USA}
}

\makeatletter
\renewcommand{\maketitle}{%
  \par
  \begingroup
    \renewcommand{\thefootnote}{\fnsymbol{footnote}}
    \setlength{\parskip}{0pt}
    \centering

    {\LARGE\scshape \@title\par}

    \vspace{12pt}

    {\normalsize
      \begin{tabular}{@{}c@{}}
        \@author
      \end{tabular}\par
    }

    \@thanks
  \endgroup
  \setcounter{footnote}{0}
  \vspace{10pt}
}
\makeatother

\makeatletter
\DeclareMathOperator{\mcirc}{%
  \mathbin{\mathpalette\on@ntimes\relax}%
}
\newcommand{\on@ntimes}[2]{%
  \vcenter{\hbox{%
    \sbox0{\m@th$#1\otimes$}%
    \setlength\unitlength{\wd0}%
    \begin{picture}(1,.8)
      \linethickness{0.35pt}
      \put(.5,.5){\circle{.9}}
    \end{picture}%
  }}%
}
\makeatother

\begin{document}

\maketitle
\begin{abstract}

Foundation models provide powerful representations for unstructured data, such as text and images, and parameter-efficient fine-tuning (PEFT) makes their adaptation to downstream tasks increasingly practical. For neural estimators that lack a tractable sampling distribution, resampling-based procedures offer a flexible route to statistical inference, but they often require hundreds of model fits, making them computationally prohibitive at foundation-model scale. We introduce HIFI, a two-stage fine-tuning framework for statistical inference on parameters such as conditional means. Stage 1 adapts a foundation model once on a larger sample, capturing the complex structure of the data and controlling estimation bias through a single fit. Stage 2 applies a resampling inference procedure to a smaller independent sample, concurrently fitting multiple lightweight PEFT estimators to the residual to quantify estimation variability without relearning the full task. HIFI accommodates a broad class of resampling procedures, including ensemble subsampling and the nonparametric bootstrap. Across vision and text experiments, HIFI scales to pretrained models ranging from tens of millions to more than ten billion parameters while providing accurate point estimates and well-calibrated confidence intervals across diverse architectures, and reduces training compute by up to 91.9\% relative to direct resampling. An application to gene-expression inference illustrates the practical value of the framework for uncertainty-aware scientific inference.

\end{abstract}

\section{Introduction}

Pretrained foundation models have changed how predictive models are built, especially for text, images, and other unstructured data \citep{bommasani2021opportunities}. Rather than estimating
every model parameter from a task-specific sample, practitioners can adapt a pretrained backbone using a comparatively small number of trainable parameters
\citep{houlsby2019parameter,li2021prefix,
lester2021power,hu2022lora,wang2025peftsurvey}. Parameter-efficient fine-tuning (PEFT) therefore makes powerful representations available when full fine-tuning is prohibitively expensive. In many scientific applications, however, an accurate point estimate is not the final goal, because reliable scientific interpretation requires quantifying the uncertainty associated with that estimate. We therefore seek a confidence interval for the population conditional mean $\mu_0(x)=\mathbb E(Y\mid X=x)$ at a fixed input $x$. 

Several approaches have been developed for inference\footnote{Throughout this paper, ``inference'' refers to statistical inference unless otherwise specified.} on conditional means estimated by neural
networks. Existing methods include approaches that exploit specialized
analytical structure
\citep{liu2022optimal,kallus2022implicit,sluijterman2025likelihood},
sample-splitting procedures such as HulC
\citep{kuchibhotla2024hulc}, and resampling methods
\citep{fei2024ulearning,meng2025esm,wu2025scalable,arie2024confidence}.
We focus on the resampling-based class because it requires relatively little
model-specific analytical structure and therefore provides a flexible route
to inference for complex neural estimators. These procedures construct uncertainty estimates from multiple fitted
estimators, but doing so introduces two central challenges. First, fitting multiple estimators does not by itself
eliminate systematic estimation error, whose control is typically required
for valid inference. Second, obtaining many sufficiently accurate fits can be
computationally expensive. This cost becomes particularly severe for large
foundation models, since each additional fit still requires substantial
forward--backward computation even when only a small subset of parameters is
updated.

\begin{figure}[t]
\vspace{-.3in}
    \centering
    \includegraphics[
        width=\textwidth
    ]{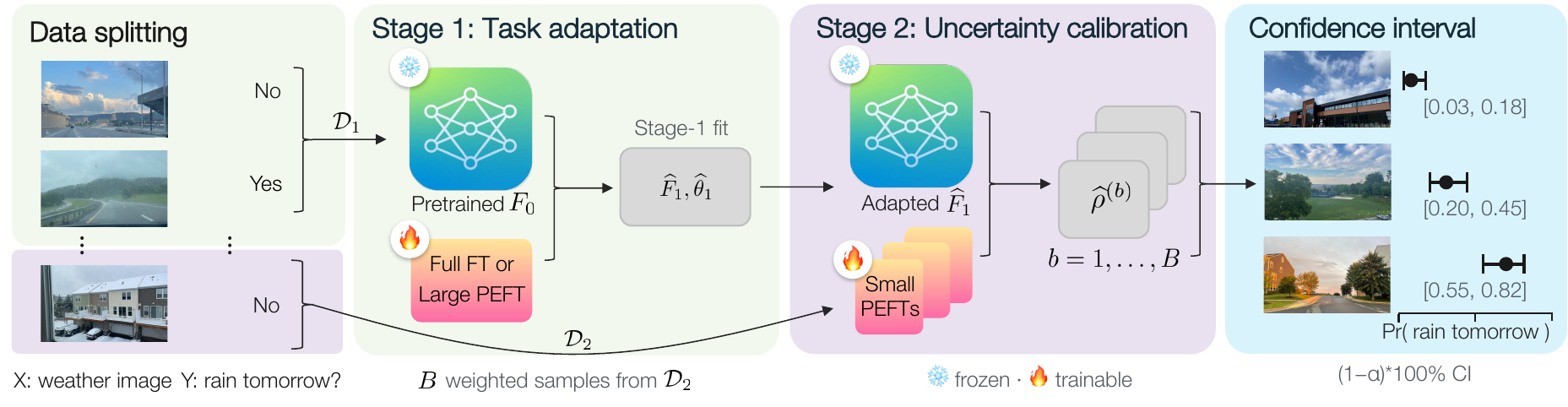}
    \caption{
Overview of HIFI. Stage~1 performs a single task-adaptation fit on
$\mathcal D_1$, yielding an adapted backbone $\widehat F_1$ and predictor
$\widehat\theta_1$. Holding these components fixed, Stage~2 fits $B$
estimator-specific residual PEFT modules using different sampling weights
on the independent sample $\mathcal D_2$. The residual estimates are then
combined to obtain a point estimate and pointwise confidence interval for
$\mu_0(x)$.
    }
    \label{fig:peft-framework}
\end{figure}

\textbf{Our contributions.}
We introduce HIFI, a highly efficient two-stage PEFT framework for statistical inference with foundation models, illustrated in Figure~\ref{fig:peft-framework}. Stage~1 uses a larger sample and a higher-capacity adaptation to capture the complex, task-relevant structure of the data, controlling estimation bias through a single fit. Stage~2 then uses a smaller independent sample to resample and ensemble lightweight PEFT estimators of the remaining residual, quantifying estimation variability without relearning the full task. Because each Stage~2 fit involves a small adaptation module on few samples, repeated fitting is less expensive, and a batched implementation trains many such estimators concurrently over the Stage~1 adapted backbone. Together, these strategies make resampling-based inference practical at foundation-model scale.

The framework accommodates multiple inference methods. We consider the
ensemble subsampling method (ESM) \citep{meng2025esm} and nonparametric
bootstrap. Across vision and text tasks and diverse pretrained architectures, the framework provides well-calibrated confidence intervals while reducing training computation by up to 92\% at the same total sample size. We further demonstrate its use in a biological application to gene-expression prediction from H\&E images.

\section{Related Work}

\textbf{Parameter-efficient adaptation.} A large literature studies parameter-efficient adaptation of pretrained models, including adapter-based, prompt- and prefix-based, and low-rank adaptation methods \citep{houlsby2019parameter,li2021prefix,lester2021power, hu2022lora,wang2025peftsurvey}. LoRA, for example, freezes the pretrained
weights and learns low-rank updates to selected weight matrices, substantially
reducing the number of trainable parameters and the memory required for
adaptation \citep{hu2022lora}. These developments make adaptation of increasingly
large pretrained models practical, but primarily target
downstream prediction rather than frequentist statistical inference.

\textbf{Uncertainty quantification for neural networks.}
Our inferential target differs from several major strands of uncertainty
quantification for modern neural networks. Deep ensembles train multiple
neural networks independently from different initializations and aggregate
their predictive distributions to quantify predictive uncertainty
\citep{lakshminarayanan2017simple}, while calibration methods assess whether
reported confidence scores reflect empirical frequencies
\citep{guo2017calibration,minderer2021revisiting}. Bayesian approaches instead
characterize uncertainty through posterior distributions over model parameters
and predictions, commonly summarized by the posterior predictive distribution
$p(y_{\mathrm{new}}\mid x,\mathcal D)
=\int p(y_{\mathrm{new}}\mid x,\phi)p(\phi\mid\mathcal D)\,d\phi$ \citep{gal2016dropout}.
Recent Bayesian PEFT methods restrict this posterior treatment to
low-dimensional adaptation parameters
\citep{yang2024bayesian,wang2024blob}. Conformal methods
construct prediction sets for future outcomes with coverage
$P\{Y_{\mathrm{new}}\in\widehat C_\alpha(X_{\mathrm{new}})\}\ge 1-\alpha$
under exchangeability
\citep{lei2018distribution,angelopoulos2023gentle}. Their inferential object is
the future response $Y_{\mathrm{new}}$, whereas our target is the population
conditional mean $\mu_0(x)=\mathbb E(Y\mid X=x)$ at a fixed input $x$, for
which we seek a confidence interval satisfying
$P\{\mu_0(x)\in C_\alpha^\mu(x)\}\to 1-\alpha$. This is often harder than predictive inference. {For example, in binary classification, an informative prediction set needs calibrated probability estimation \citep{guo2017calibration}, whereas a confidence interval for $\mu_0(x)$ must shrink as data accumulate, and its validity requires the estimation bias to be negligible relative to the estimation variability, beyond calibration.} 

\textbf{Frequentist inference for neural networks.}
Several approaches have been developed for frequentist inference on
neural-network estimators, including specially constructed estimators
\citep{liu2022optimal}, objective perturbation
\citep{kallus2022implicit}, and likelihood-ratio methods
\citep{sluijterman2025likelihood}. These approaches provide principled
inferential guarantees but rely on analytical structure specific to the estimator or training objective. A different route is provided by HulC, which constructs confidence sets from estimators trained on independent data splits, with validity governed by their median bias \citep{kuchibhotla2024hulc}. In parallel, much of the literature on
frequentist inference for neural-network regression relies on resampling-based methods, which require relatively little model-specific analytical structure and can therefore be applied more flexibly to complex neural estimators. Ensemble-subsampling approaches train estimators on multiple subsamples and exploit their dependence structure for inference \citep{fei2024ulearning,meng2025esm,wu2025scalable}, while nonparametric bootstrap methods refit neural estimators on resampled datasets \citep{arie2024confidence}. However, these methods have primarily been developed with conventional DNNs,
with computational savings obtained through parallelization or smaller
subsamples rather than batched multi-estimator training.

\textbf{Efficient ensemble and multi-adapter training.}
A separate line of work develops efficient mechanisms for training multiple
predictors or parameter-efficient adapters over a shared backbone.
BatchEnsemble shares model parameters across ensemble members and updates
multiple members concurrently \citep{wen2020batchensemble}, while LoRA-based
ensemble methods use a common pretrained backbone with member-specific
low-rank adapters for predictive uncertainty and calibration
\citep{wang2023loraensemble,muehlematter2026loraensemble}. More recently,
\citet{li2026tlora} develop systems support for concurrently training
multiple LoRA jobs over a shared backbone. These methods primarily target efficient ensembling, predictive uncertainty,
or multi-adapter training. Our work instead brings these computational ideas
into resampling-based frequentist inference by using batched PEFT training.

\section{HIFI Framework}
\label{sec:framework}

Let $Z=(X,Y)\sim P$, where $X$ is a potentially high-dimensional or
unstructured input and $Y$ is a scalar response. The inferential target is
the conditional mean $\mu_0(x)=\mathbb E(Y\mid X=x)$. We work on the link
scale $\theta_0(x)=g\{\mu_0(x)\}$, where $g$ is a known monotone link
function, including the identity, logit, and log links for continuous,
binary, and count outcomes, respectively. Pointwise inference is performed
for $\theta_0(x)$ and transformed to the response scale through $g^{-1}$.

As illustrated in Figure~\ref{fig:peft-framework}, we separate task
adaptation from statistical inference through a two-stage sample-splitting strategy. We split the labeled data into two independent samples,
$\mathcal D_1=\{Z_i^{(1)}\}_{i=1}^{n_1}$ and
$\mathcal D_2=\{Z_i^{(2)}\}_{i=1}^{n_2}$.
The first sample is used for a single Stage-1 task adaptation,
while the second is used by resampling-based procedures to construct Stage-2
estimators for statistical inference.

\subsection{Stage-1 Task Adaptation}
\label{sec:first_stage}
Let $F_0$ denote a pretrained foundation model. Stage~1 applies an
adaptation procedure $\mathcal A_1$ to $F_0$ using $\mathcal D_1$,
yielding $(\widehat F_1,\widehat\theta_1)
=\mathcal A_1(F_0,\mathcal D_1,U_1)$,
where $\widehat F_1$ is the adapted backbone,
$\widehat\theta_1$ is the corresponding predictor with a task-specific
prediction head, and $U_1$ collects training randomness. Stage~1 may
use full fine-tuning or a higher-capacity PEFT parameterization.

To specify the Stage-1 optimization, let $\ell_1(y,t)$ denote a loss, whose conditional population
risk is minimized at $t=\theta_0(x)$. The training objective is
\begin{equation}
    \widehat{\mathcal L}_1(\theta)
    =
    \frac{1}{n_1}
    \sum_{i=1}^{n_1}
    \ell_1\left(
        Y_i^{(1)},
        \theta(X_i^{(1)})
    \right),
    \label{eq:stage1_objective}
\end{equation}
where $\theta$ ranges over predictors induced by the adaptation
parameterization and prediction head.

\subsection{Stage 2: Uncertainty Calibration}
\label{sec:stage2}

Conditional on the Stage-1 fit, define the residual signal as
$\rho_0(x)=\theta_0(x)-\widehat\theta_1(x)$.
Stage~2 estimates it using $B$ sampled or weighted
versions of $\mathcal D_2$, represented by sampling-weight vectors
$W_1,\ldots,W_B$, with $W_b=(W_{b1},\ldots,W_{bn_2})$.
The sampling weights depend on the inference method
$\mathfrak I_\alpha$.

For each $b$, the Stage-1 adapted backbone $\widehat F_1$ and
predictor $\widehat\theta_1$ are held fixed and shared across all
Stage~2 fits. Starting from $\widehat F_1$, the $b$th fit learns
estimator-specific PEFT parameters and a residual prediction head,
yielding
$\widehat\rho_b
=\mathcal A_2(\widehat F_1,\mathcal D_2,W_b;
\widehat\theta_1,U_{2,b})$,
where $\mathcal A_2$ denotes the Stage-2 adaptation procedure and
$U_{2,b}$ collects training randomness.

More specifically, let $\ell_2(y,t)$ denote the Stage-2 loss, which may be the same as or different from the Stage-1 loss $\ell_1$, and whose conditional population risk is minimized at $t=\theta_0(x)$. Holding $\widehat\theta_1$ fixed, the $b$th training objective is
\begin{equation}
    \widehat{\mathcal L}_{2,b}(\rho)
    =
    \frac{1}{\sum_{i=1}^{n_2}W_{bi}}
    \sum_{i=1}^{n_2}
    W_{bi}\,
    \ell_2\!\left(
        Y_i^{(2)},
        \widehat\theta_1(X_i^{(2)})
        +
        \rho(X_i^{(2)})
    \right).
    \label{eq:stage2_objective}
\end{equation}
Let $C_\alpha^\rho(x)$ denote a $(1-\alpha)$ confidence interval for
$\rho_0(x)$. An inference method $\mathfrak I_\alpha$ constructs it from
the sampling weights and Stage-2 residual estimates as
$C_\alpha^\rho(x)
=\mathfrak I_\alpha(\{W_b,\widehat\rho_b(x)\}_{b=1}^B)$.
Since $\theta_0(x)=\widehat\theta_1(x)+\rho_0(x)$ conditional on
Stage 1, the corresponding intervals on the link and response scales are
$C_\alpha^\theta(x)=\widehat\theta_1(x)+C_\alpha^\rho(x)$ and
$C_\alpha^\mu(x)=g^{-1}\{C_\alpha^\theta(x)\}$, respectively.
This formulation is method-agnostic. In this paper, we use ESM and bootstrap.

\subsection{HIFI Instantiation with LoRA and ESM}
\label{sec:instantiation}

We illustrate HIFI using LoRA with ESM. The corresponding bootstrap instantiation is given in Appendix~\ref{app:bootstrap_implementation}. In Stage~1, for each selected pretrained weight matrix $M_{\ell,0}$,
LoRA learns
\[
    \widehat M_{\ell,1}
    =
    M_{\ell,0}
    +
    \frac{\lambda_1}{q_1}
    \widehat L_{\ell,1}\widehat R_{\ell,1},
\]
where $q_1$ and $\lambda_1$ are the LoRA rank and scaling parameter, respectively.
The LoRA factors and prediction head are optimized according to
\eqref{eq:stage1_objective}, yielding the adapted backbone
$\widehat F_1$ and predictor $\widehat\theta_1$.

For Stage~2, ESM independently draws $B$ size-$r$ subsets
$I_1,\ldots,I_B$ of $\{1,\ldots,n_2\}$, with
$W_{bi}=\mathbf 1\{i\in I_b\}$. Holding the Stage~1 adapted matrices
fixed, the $b$th fit uses
\[
    \widehat M_{\ell,2,b}
    =
    \widehat M_{\ell,1}
    +
    \frac{\lambda_2}{q_2}
    \widehat L_{\ell,2,b}\widehat R_{\ell,2,b},
    \qquad b=1,\ldots,B,
\]
where $q_2$ and $\lambda_2$ are the Stage~2 LoRA rank and scaling
parameter, respectively. Each fit learns new LoRA factors and a residual prediction
head according to \eqref{eq:stage2_objective}, yielding
$\widehat\rho_b$.

ESM averages the residual estimates as
$\widehat\rho_{\mathrm{ESM}}(x)
=B^{-1}\sum_{b=1}^B\widehat\rho_b(x)$, yielding the link-scale point
estimate
$\widehat\theta_{\mathrm{ESM}}(x)
=\widehat\theta_1(x)+\widehat\rho_{\mathrm{ESM}}(x)$.
Let $\overline W_i=B^{-1}\sum_{b=1}^B W_{bi}$. The infinitesimal
jackknife variance estimator is
\[
    \widehat s_{\mathrm{ESM}}^2(x)
    =
    \frac{n_2(n_2-1)}{(n_2-r)^2}
    \sum_{i=1}^{n_2}
    \left[
        \frac{1}{B}\sum_{b=1}^B
        (W_{bi}-\overline W_i)
        \{\widehat\rho_b(x)-\widehat\rho_{\mathrm{ESM}}(x)\}
    \right]^2.
\]
The resulting pointwise interval is
$C_{\alpha}^{\theta,\mathrm{ESM}}(x)
=
\widehat\theta_{\mathrm{ESM}}(x)
\pm z_{1-\alpha/2}\widehat s_{\mathrm{ESM}}(x)$,
with the response-scale interval obtained through $g^{-1}$ and point estimate
$\widehat\mu_{\mathrm{ESM}}(x)
=
g^{-1}\{\widehat\theta_{\mathrm{ESM}}(x)\}$.

\subsection{Efficient Batched Stage-2 Training}
\label{sec:implementation}

The resampling-based Stage~2 procedures above require fitting $B$ estimators, which can be computationally expensive when performed sequentially even under PEFT. To reduce this overhead, we stack the estimator-specific LoRA parameters and prediction heads along an additional ensemble dimension and execute their forward and backward computations using batched tensor
operations on the GPU. The frozen backbone weights are shared across fits. For a group of $B_{\mathrm{chunk}}$ fits, each fit retains its own resampling weights, minibatch, loss, and optimizer state, while gradient updates are confined to its estimator-specific parameter slice. This batching strategy improves GPU utilization by replacing separate training loops with larger batched operations and is algebraically equivalent to fitting the same estimators separately, up to numerical differences.

The $B$ fits are processed in groups of at most
$B_{\mathrm{chunk}}$ estimators, with group size determined by GPU
memory. Predictions are collected across groups before constructing
confidence intervals. Although we use ESM and bootstrap, the same construction applies to other resampling-based procedures.

\section{Theoretical Motivation}
\label{sec:inferential_properties}

We use an idealized sparse ReLU-DNN setting to motivate the
statistical and computational rationale of HIFI. Building on
approximation--estimation risk decompositions
\citep{schmidt2020nonparametric,meng2025esm},
Appendix~\ref{app:hifi_theory} derives risk bounds under explicit
compatibility and learning assumptions. We then describe additional
conditions under which conditional ESM validity transfers to
unconditional coverage. This analysis provides theoretical motivation
for the two-stage design.

Consider the $\gamma$-H\"older, $d$-dimensional special case of the
bounded sparse ReLU-DNN setting in \citet{meng2025esm}. Let
$\widehat\theta^{\mathrm{single}}$ and
$\widehat\theta^{\mathrm{HIFI}}$ denote the direct-resampling and
HIFI estimators, respectively, and let
$\nu=2\gamma/(2\gamma+d)$. Let $r$ be the Stage~2 subsample size,
and suppose the direct-resampling subsample size and Stage~1 sample
size are $c_0r$ and $c_1r$, respectively, for fixed
$c_0,c_1\ge1$. Applying
Proposition~\ref{prop:hifi_risk_explicit} in
Appendix~\ref{app:hifi_learning_setup} together with the
corresponding DNN excess-risk rate gives
\begin{equation}
\begin{aligned}
\mathbb E\|\widehat\theta^{\mathrm{single}}-\theta_0\|_2^2
&\lesssim
(c_0r)^{-\nu}L_{c_0r}\log^2(c_0r),\\
\mathbb E\|\widehat\theta^{\mathrm{HIFI}}-\theta_0\|_2^2
&\lesssim
(c_1r)^{-\nu}L_{c_1r}\log^2(c_1r)
+
r^{-\nu}L_r\log^2 r.
\end{aligned}
\label{eq:hifi_main_risk}
\end{equation}
Here $L_s$ denotes the network depth for a fit on $s$ observations.

The key compatibility requirement is that, conditional on Stage~1, the remaining Stage~2 class retains sufficient approximation capacity for the residual target
(Assumption~\ref{ass:hifi_stage1_learning} in
Appendix~\ref{app:hifi_learning_setup}).
Single-stage resampling refits all parameters on each subsample of size $c_0r$, whereas HIFI fits the shared component once using $c_1r$ observations and refits only the Stage~2 component using subsamples of size $r$. For comparable depth scaling, the two bounds have the same asymptotic rate order in $r$. Thus, HIFI reduces repeated fitting without changing the nonparametric rate order.

\textbf{Valid inference.}
\label{sec:common_validity}
Write $h=(F,t)$ for a generic Stage~1 output and
$\widehat h_1=(\widehat F_1,\widehat\theta_1)$ for the fitted Stage~1 output. For fixed $h$, define the residual target $\rho_h=\theta_0-t$. In addition to the excess-risk control in \eqref{eq:hifi_main_risk}, valid inference requires the regularity conditions of the chosen resampling method. Conditional on a deterministic admissible $h$, Stage~2 targets the fixed residual function $\rho_h$. Suppose that, with $h$ held fixed, the resulting Stage~2 DNN--ESM problem satisfies the conditions in Corollary~3.6 and Theorems~3.7--3.8 of \citet{meng2025esm}. Then the corresponding conditional inference is asymptotically valid.

To account for Stage~1 randomness, suppose there exist deterministic sets $\mathcal H_{n_2}$ and $\mathcal T_{n_2}\subseteq\mathcal X$ with $P(\widehat h_1\in\mathcal H_{n_2})\to1$ and $P_X(\mathcal T_{n_2})\to1$, such that conditional ESM validity holds uniformly over $h\in\mathcal H_{n_2}$ and $x\in\mathcal T_{n_2}$. Then, for every deterministic sequence $x_{n_2}\in\mathcal T_{n_2}$, averaging over Stage~1 randomness yields $P\{\theta_0(x_{n_2})\in C_{\alpha}^{\theta,\mathrm{ESM}}(x_{n_2};\widehat h_1)\}\to1-\alpha$. Thus, the HIFI--ESM interval attains asymptotically nominal unconditional coverage at any deterministic target sequence lying in the asymptotically full-probability target sets $\mathcal T_{n_2}$. Details are given in Appendix~\ref{app:hifi_pointwise_control}.

\textbf{Computational gains.}
\label{sec:hifi_allocation}
Suppose training cost scales as the product of trainable parameters
and sample size. With $B$ resampling fits, single-stage resampling
costs $\mathcal C_{\mathrm{single}}=Bp_0c_0r$, whereas HIFI costs
$\mathcal C_{\mathrm{HIFI}}=p_1c_1r+Bp_2r$, since the larger
Stage~1 component is trained only once. For $p_1=0.8p_0$,
$p_2=0.2p_0$, $c_0=4$, and $c_1=16$, corresponding to an
80/20 allocation of trainable parameters between Stages~1 and~2
and sample sizes $4r$ and $16r$ for the single-stage and Stage~1
fits, respectively, \eqref{eq:hifi_main_risk} gives the same
excess-risk order under comparable depth scaling, while
$\mathcal C_{\mathrm{HIFI}}/\mathcal C_{\mathrm{single}}
=3.2/B+0.05$. At $B=500$, HIFI therefore uses $5.64\%$ of the
training cost.

\section{Synthetic Experiments}

We evaluate HIFI across two modalities and six pretrained backbones. The vision task considers continuous-outcome regression, whereas the text task considers Bernoulli probability estimation. All methods use LoRA, with HIFI applying it in both stages. We evaluate ESM and bootstrap within HIFI and compare against the external comparators HulC \citep{kuchibhotla2024hulc} and Bayesian-PEFT, adapted from the Bayesian neural-network framework of \citet{sun2021sparse}.

Unless otherwise stated, each experiment uses 100 replications with independently generated Stage~1 and Stage~2 samples, split into 8,000/500 and 1,000/200 training/validation observations, respectively. Training in both stages uses early stopping based on the corresponding validation set. For the external comparators, the Stage~1 and Stage~2 samples are pooled, so that each method uses all 9,700 observations. ESM uses 500 subsamples of size 501 without replacement, while bootstrap uses 400 resamples of size 1,000 with replacement. For evaluation, the vision task uses 500 target images and the text task uses 424 target texts, all held constant across replications. Implementation details for the inference methods and external comparators are provided in
Appendix~\ref{app:inference-implementation}, with model-specific
architecture, adaptation, and optimization settings reported in
Appendix~\ref{app:experiment-details}.

\subsection{Task 1: Regression from Synthetic Images}
\label{sec:vision-task}

We consider a vision-regression setting in which each input $X$ is
a $224\times224$ RGB image containing a single geometric object.
The shape $S$ is sampled uniformly from $\{\mcirc,\Box,\triangle\}$, and its
geometric area fraction $a$ is sampled from
$\mathrm{Unif}(0.10,0.32)$.
Distinct foreground and background colors are sampled from a
four-color palette, while object locations and the orientations of
squares and triangles are randomized. Responses are generated as
$Y=\beta(S)a+\varepsilon$, where $\varepsilon\sim N(0,0.25^2)$,
$\beta(\mcirc)=1$, $\beta(\Box)=2$, and
$\beta(\triangle)=3$.
For a fixed target image $x$, let $S(x)$ and $a(x)$ denote its
generating shape and geometric area fraction. By construction,
$\mu_0(x)=\beta\{S(x)\}a(x)$, and our goal is to estimate
$\mu_0(x)$ and construct a pointwise confidence interval. The model is trained only on the observed
image--response pairs $(X,Y)$ and is not given $S(X)$ or $a(X)$.
Additional data-generation details are provided in
Appendix~\ref{app:vision-dgp}, with representative images
shown in Figure~\ref{fig:vit-simulation-examples}.

We evaluate three pretrained vision backbones spanning different architectural and pretraining designs: ViT-B/32
\citep{dosovitskiy2021an}, a standard vision transformer;
DINOv2 ViT-B/14 \citep{oquab2024dinov}, a self-supervised
vision transformer; and Swin-B \citep{liu2021swin}, a hierarchical shifted-window transformer. Across all three pretrained vision backbones, HIFI achieves low point-estimation error and stable near-nominal coverage. At the 95\% level, HIFI-ESM achieves coverage of 0.937--0.955 and HIFI-Bootstrap 0.938--0.953 (Table~\ref{tab:vision_main_results}), with HIFI-Bootstrap yielding the shorter 95\% intervals across all three backbones. In contrast, HulC and Bayesian-PEFT exhibit less stable coverage across backbones, with 95\% coverage ranging from 0.906 to 0.965 and 0.764 to 0.993, respectively. For 95\% coverage, target-wise standard deviations remain within 0.03--0.06 for HIFI, compared with 0.02--0.12 for HulC and 0.03--0.32 for Bayesian-PEFT across backbones, while additional target-level diagnostics in Appendix~\ref{app:vision-target-diagnostics} show similarly stable behavior, indicating that HIFI provides robust inference across targets within each architecture.

\begin{table}[t]
\vspace{-.3in}
\caption{
Statistical inference with HIFI and external baselines across pretrained vision models.
}
\centering
\label{tab:vision_main_results}
\footnotesize
\setlength{\tabcolsep}{1.8pt}
\renewcommand{\arraystretch}{0.95}

\begin{tabular}{@{}lll cc ccc@{}}
\toprule
& & &
\multicolumn{2}{c}{Point Estimation} &
\multicolumn{3}{c}{Statistical Inference} \\
\cmidrule(lr){4-5}
\cmidrule(lr){6-8}
Model
& Params
& Method
& MAB
& MAE
& Cov.\ 90\%
& Cov.\ 95\%
& 95\% CI Length \\
\midrule

\rowcolor{hifishade}
\cellcolor{white} & \cellcolor{white} & \textsc{HIFI}-ESM
& 0.018 (0.02)
& 0.041 (0.01)
& 0.908 (0.08)
& 0.937 (0.06)
& 0.282 (0.02) \\

\rowcolor{hifishade}
\cellcolor{white} & \cellcolor{white} & \textsc{HIFI}-Bootstrap
& 0.017 (0.02)
& 0.047 (0.01)
& 0.888 (0.07)
& 0.938 (0.05)
& 0.231 (0.02) \\

\cmidrule(lr){3-8}

& & HulC
& 0.028 (0.03)
& 0.040 (0.02)
& 0.867 (0.11)
& 0.914 (0.10)
& 0.206 (0.03) \\

\multirow{-4}{*}{ViT-B/32}
& \multirow{-4}{*}{87.5M}
& Bayesian-PEFT
& 0.054 (0.05)
& 0.060 (0.05)
& 0.702 (0.33)
& 0.764 (0.32)
& 0.183 (0.04) \\

\midrule

\rowcolor{hifishade}
\cellcolor{white} & \cellcolor{white} & \textsc{HIFI}-ESM
& 0.011 (0.01)
& 0.044 (0.01)
& 0.933 (0.04)
& 0.955 (0.03)
& 0.319 (0.04) \\

\rowcolor{hifishade}
\cellcolor{white} & \cellcolor{white} & \textsc{HIFI}-Bootstrap
& 0.011 (0.01)
& 0.046 (0.01)
& 0.908 (0.04)
& 0.953 (0.03)
& 0.243 (0.03) \\

\cmidrule(lr){3-8}

& & HulC
& 0.021 (0.02)
& 0.073 (0.02)
& 0.931 (0.03)
& 0.965 (0.02)
& 0.498 (0.13) \\

\multirow{-4}{*}{DINOv2-B/14}
& \multirow{-4}{*}{86.6M}
& Bayesian-PEFT
& 0.029 (0.03)
& 0.049 (0.03)
& 0.984 (0.05)
& 0.993 (0.03)
& 0.484 (0.04) \\

\midrule

\rowcolor{hifishade}
\cellcolor{white} & \cellcolor{white} & \textsc{HIFI}-ESM
& 0.013 (0.01)
& 0.035 (0.01)
& 0.925 (0.05)
& 0.951 (0.04)
& 0.236 (0.03) \\

\rowcolor{hifishade}
\cellcolor{white} & \cellcolor{white} & \textsc{HIFI}-Bootstrap
& 0.013 (0.01)
& 0.042 (0.01)
& 0.893 (0.06)
& 0.943 (0.04)
& 0.215 (0.03) \\

\cmidrule(lr){3-8}

& & HulC
& 0.028 (0.03)
& 0.041 (0.02)
& 0.860 (0.13)
& 0.906 (0.12)
& 0.207 (0.03) \\

\multirow{-4}{*}{Swin-B}
& \multirow{-4}{*}{87.8M}
& Bayesian-PEFT
& 0.033 (0.03)
& 0.057 (0.03)
& 0.881 (0.15)
& 0.925 (0.12)
& 0.281 (0.03) \\

\bottomrule
\end{tabular}

\vspace{2pt}
\begin{minipage}{0.98\textwidth}
\footnotesize
\textit{Notes.}
MAB/MAE denote mean absolute bias/error; parentheses report target-wise SD; Bayesian-PEFT reports posterior credible intervals.
\end{minipage}
\end{table}

\subsection{Task 2: Probability Prediction from Text}
\label{sec:text-task}

We consider a controlled bilingual text task in which each input $X$
describes when a passenger arrives relative to a scheduled flight, and
$Y=1$ indicates that the passenger catches the flight.
Let $T(X)\in[0,3.5]$ denote the arrival lead time encoded in the text,
measured in hours. Larger values of
$T(X)$ correspond to earlier arrivals and therefore higher success
probabilities. Specifically, the conditional success probability is
$\mu_0(X)=\operatorname{logit}^{-1}\{2.8[T(X)-1.05]\}$, and
$Y\mid X\sim\operatorname{Bernoulli}\{\mu_0(X)\}$.
For a fixed target text $x$, our goal is to estimate $\mu_0(x)$ and
construct a pointwise confidence interval.

For each observation, the arrival lead time $T$ is generated using probability-stratified sampling, and the language is independently drawn as English or Chinese with equal probability. Conditional on $T$ and the sampled language, the arrival-time information is expressed in natural language to form the text input $X$. Additional auxiliary flight information and response-irrelevant contextual information are included randomly to introduce nuisance textual variation. During training, the model observes only $(X,Y)$ and is not given $T(X)$ or $\mu_0(X)$ explicitly. Representative examples are shown in Appendix Table~\ref{tab:text-dgp-examples}, with complete data-generation details provided in Appendix~\ref{app:text-dgp}.

We evaluate three multi-billion-parameter language backbones with distinct designs: Qwen3-Embedding-8B \citep{zhang2025qwen3}, a decoder-only embedding model; XLM-R XXL \citep{goyal2021larger}, an encoder-only multilingual masked language model; and NV-Embed-v2 \citep{lee2025nv}, a Mistral-based embedding model
with latent-attention pooling.

The language experiments demonstrate that HIFI makes resampling-based inference feasible with multi-billion-parameter foundation models while maintaining reliable statistical performance. Across three heterogeneous language backbones, HIFI-ESM attains
95\% coverage of 0.926--0.945, while HIFI-Bootstrap ranges from 0.930--0.970 (Table~\ref{tab:text_main_results}). By comparison,
the external baselines show greater variation across architectures, with coverage of
0.885--0.932 for HulC and 0.598--0.998 for Bayesian-PEFT. HIFI also exhibits stable target-level calibration, with 95\% coverage standard deviations of
0.02--0.05 across backbones, compared with 0.06--0.10 for HulC and 0.01--0.45 for Bayesian-PEFT. Additional analyses in Appendix~\ref{app:text-target-diagnostics} show similarly stable behavior across targets.

\begin{table}[t]
\vspace{-.3in}
\caption{
Statistical inference with HIFI and external baselines across pretrained language models.
}
\centering
\label{tab:text_main_results}
\footnotesize
\setlength{\tabcolsep}{1.8pt}
\renewcommand{\arraystretch}{0.95}

\begin{tabular}{@{}lll
                cc
                ccc@{}}
\toprule
& & &
\multicolumn{2}{c}{Point Estimation} &
\multicolumn{3}{c}{Statistical Inference} \\
\cmidrule(lr){4-5}
\cmidrule(lr){6-8}
Model
& Params
& Method
& MAB
& MAE
& Cov.\ 90\%
& Cov.\ 95\%
& 95\% CI Length \\
\midrule

\multirow{4}{*}{\makecell[l]{Qwen3-\\Embedding-8B}}
& \multirow{4}{*}{7.6B}
& \cellcolor{hifishade}\textsc{HIFI}-ESM
& \cellcolor{hifishade}0.014 (0.01)
& \cellcolor{hifishade}0.060 (0.01)
& \cellcolor{hifishade}0.899 (0.03)
& \cellcolor{hifishade}0.928 (0.03)
& \cellcolor{hifishade}0.374 (0.06) \\

&
& \cellcolor{hifishade}\textsc{HIFI}-Bootstrap
& \cellcolor{hifishade}0.015 (0.01)
& \cellcolor{hifishade}0.064 (0.02)
& \cellcolor{hifishade}0.885 (0.04)
& \cellcolor{hifishade}0.937 (0.03)
& \cellcolor{hifishade}0.316 (0.07) \\

\cmidrule(lr){3-8}

&
& HulC
& 0.075 (0.05)
& 0.097 (0.04)
& 0.849 (0.10)
& 0.889 (0.09)
& 0.418 (0.06) \\

&
& Bayesian-PEFT
& 0.085 (0.06)
& 0.102 (0.05)
& 0.830 (0.22)
& 0.890 (0.18)
& 0.426 (0.05) \\

\midrule

\multirow{4}{*}{XLM-R XXL}
& \multirow{4}{*}{10.7B}
& \cellcolor{hifishade}\textsc{HIFI}-ESM
& \cellcolor{hifishade}0.014 (0.01)
& \cellcolor{hifishade}0.051 (0.01)
& \cellcolor{hifishade}0.899 (0.04)
& \cellcolor{hifishade}0.926 (0.03)
& \cellcolor{hifishade}0.361 (0.08) \\

&
& \cellcolor{hifishade}\textsc{HIFI}-Bootstrap
& \cellcolor{hifishade}0.016 (0.01)
& \cellcolor{hifishade}0.061 (0.02)
& \cellcolor{hifishade}0.878 (0.07)
& \cellcolor{hifishade}0.930 (0.05)
& \cellcolor{hifishade}0.311 (0.10) \\

\cmidrule(lr){3-8}

&
& HulC
& 0.035 (0.03)
& 0.057 (0.02)
& 0.882 (0.07)
& 0.932 (0.06)
& 0.313 (0.06) \\

&
& Bayesian-PEFT
& 0.031 (0.02)
& 0.042 (0.02)
& 0.993 (0.03)
& 0.998 (0.01)
& 0.359 (0.05) \\

\midrule

\multirow{4}{*}{NV-Embed-v2}
& \multirow{4}{*}{7.9B}
& \cellcolor{hifishade}\textsc{HIFI}-ESM
& \cellcolor{hifishade}0.013 (0.01)
& \cellcolor{hifishade}0.052 (0.01)
& \cellcolor{hifishade}0.935 (0.02)
& \cellcolor{hifishade}0.945 (0.02)
& \cellcolor{hifishade}0.522 (0.08) \\

&
& \cellcolor{hifishade}\textsc{HIFI}-Bootstrap
& \cellcolor{hifishade}0.010 (0.01)
& \cellcolor{hifishade}0.048 (0.01)
& \cellcolor{hifishade}0.952 (0.03)
& \cellcolor{hifishade}0.970 (0.02)
& \cellcolor{hifishade}0.321 (0.08) \\

\cmidrule(lr){3-8}

&
& HulC
& 0.052 (0.04)
& 0.069 (0.03)
& 0.861 (0.11)
& 0.885 (0.10)
& 0.306 (0.07) \\

&
& Bayesian-PEFT
& 0.099 (0.07)
& 0.101 (0.07)
& 0.573 (0.45)
& 0.598 (0.45)
& 0.266 (0.02) \\

\bottomrule
\end{tabular}

\vspace{2pt}
\begin{minipage}{0.98\textwidth}
\footnotesize
\textit{Notes.}
MAB/MAE denote mean absolute bias/error; parentheses report target-wise SD. Bayesian-PEFT reports posterior credible intervals.
\end{minipage}
\end{table}

\subsection{Ablation Study 1: Effect of Stage-1 Task-Adaptation}
\label{sec:ablation-stage1-size}

We study the effect of adding Stage~1 task-adaptation data while holding the Stage~2 inference sample fixed. Using the bilingual flight simulation with Qwen3-Embedding-8B, we compare single-stage direct-resampling baselines based on ESM and bootstrap, which operate directly on the Stage~2-sized sample, with HIFI using $n_1\in\{2{,}000,4{,}500,8{,}500\}$ additional Stage~1 observations. Except for the Stage~1 sample size, all simulation settings, evaluation targets, training settings, and inference settings are kept identical to those of the main experiment and Appendix Table~\ref{tab:text_model_settings}.

\begin{figure}[t]
\vspace{-.2in}
\centering
\includegraphics[width=.9\textwidth]{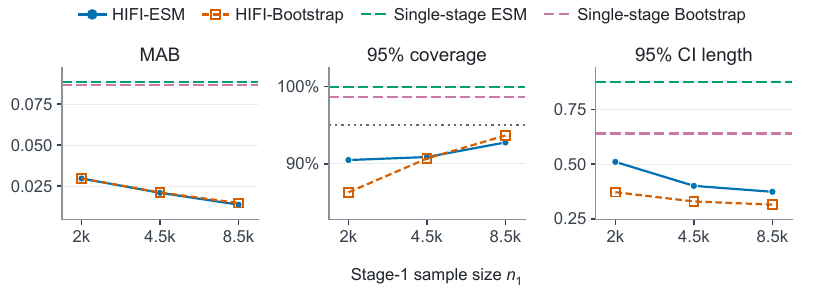}
\vspace{-.1in}
\caption{Effect of Stage~1 task adaptation on MAB,
95\% coverage, and CI length.}
\label{fig:ablation-s1-main}
\end{figure}

Figure~\ref{fig:ablation-s1-main} shows two patterns. First, augmenting the single-stage procedures with a Stage~1 task-adaptation sample substantially improves estimation accuracy and produces much shorter confidence intervals. The single-stage baselines attain conservative coverage but exhibit substantially larger estimation bias. Second, within HIFI, increasing the Stage~1 sample size further improves performance. For both ESM and bootstrap, MAB decreases and 95\% confidence intervals narrow steadily as $n_1$ increases, while coverage moves toward the nominal 95\% level. At $n_1=8{,}500$, mean interval lengths are 51--57\% smaller than those of the corresponding single-stage baselines, while empirical coverage reaches 92.8--93.7\%, slightly below nominal. Appendix~\ref{app:ablation-stage1-size} additionally reports the 90\% results and target-wise analyses of estimation, coverage, and confidence-interval length, showing that the aggregate patterns are broadly consistent across targets.

\subsection{Ablation Study 2: HIFI vs. Direct Resampling at Matched Sample Size}

We further compare HIFI with direct single-stage resampling under a matched total sample size. The single-stage baseline pools the two samples (9,000 training and 700 validation observations) and applies ESM or bootstrap directly, using the same PEFT specification as HIFI
Stage~2. We evaluate the three vision backbones and
Qwen3-Embedding-8B with all other settings unchanged.

\begin{figure}[t]
    \centering
    \vspace{-.1in}\includegraphics[width=.9\linewidth]{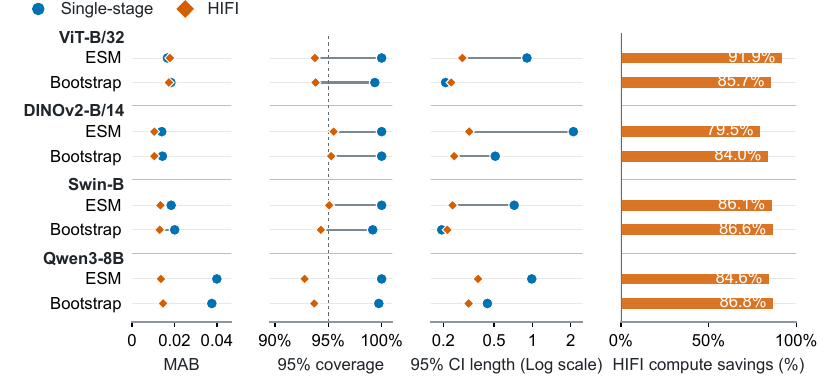}
    \vspace{-.1in}
    \caption{
 HIFI versus direct single-stage resampling under the same total data budget. HIFI training-compute savings measured in EFLOPs.
    }
    \label{fig:ablation-direct}
\end{figure}

Figure~\ref{fig:ablation-direct} shows that the two-stage framework improves the statistical--computational trade-off at the same total sample size. Direct resampling is strongly conservative, with 95\% coverage concentrated near 100\% across model and inference methods, whereas HIFI tracks the nominal level much more
closely, with coverage ranging from 92.8\% to 95.5\%.
For ESM, HIFI also produces substantially shorter intervals across all four backbones. For bootstrap, intervals are shorter for DINOv2-B/14 and
Qwen3-Embedding-8B and comparable for the other backbones. HIFI also yields lower MAB in seven of the eight comparisons with the largest improvements for Qwen3-Embedding-8B.

These statistical gains are accompanied by substantial computational savings: HIFI reduces training EFLOPs by 79.5--91.9\% relative to direct resampling. Detailed training-compute calculations and wall-clock times are reported in Appendix~\ref{app:computation}. The results are consistent with the two-stage design amortizing task
adaptation in Stage~1 and using Stage~2 for uncertainty calibration. Additional 90\% and target-wise results in
Appendix~\ref{app:ablation_direct} show the same pattern.

\section{Real Data Analysis}
\label{sec:real_data}

Spatial transcriptomics provides spatially resolved but inherently noisy measurements of gene expression. Estimating the conditional mean $\mu_0(x)=\mathbb E(Y\mid X=x)$ from local H\&E morphology denoises these measurements and enriches the underlying signal \citep{zhang2024inferring}, and the resulting expression maps are routinely used in downstream analyses. Without uncertainty quantification, however, estimation error can be mistaken for biological structure and propagate into these analyses.

To demonstrate the practical value of HIFI, we use the Xenium human breast cancer dataset from \citet{janesick2023high}, which contains two tissue sections. We train on Section~1 and evaluate on the held-out Section~2, using its H\&E image to predict all 313 measured genes, with the observed expression used for evaluation. To facilitate
a fair comparison with iStar \citep{zhang2024inferring}, we adopt the
same preprocessing pipeline and HViT backbone, while using our HIFI
framework for model fitting. Additional implementation details and
experimental settings are provided in Appendix~\ref{app:xenium}.

\begin{figure}[t]
    \centering
    \vspace{-.2in}\includegraphics[width=0.99\linewidth]{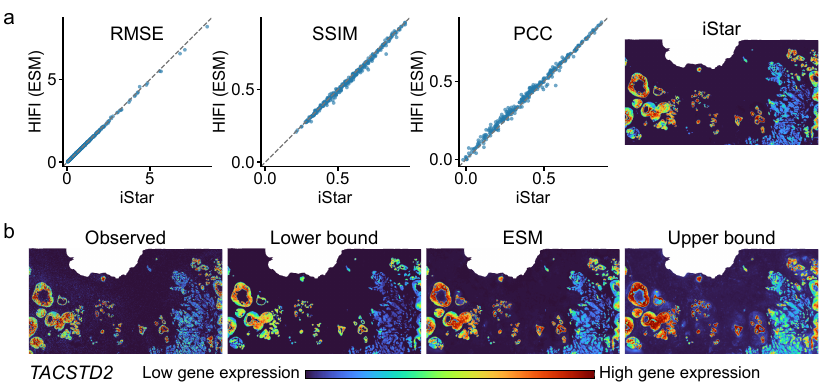}
    \vspace{-.1in}
\caption{
\textbf{Gene-expression inference on Xenium tissue Section~2.}
\textbf{(a)} HIFI with ESM versus iStar across 313 genes by RMSE, SSIM,
and PCC, with the iStar point estimate for \textit{TACSTD2}.
\textbf{(b)} HIFI point estimate and 95\% pointwise confidence interval
for \textit{TACSTD2}.
}
    \label{fig:xenium-main}
\end{figure}

HIFI retains point-prediction performance comparable to iStar while additionally enabling uncertainty quantification. As shown in
Figure~\ref{fig:xenium-main}a, HIFI with ESM performs comparably to
iStar across root mean squared error (RMSE), structural similarity
index (SSIM), and Pearson correlation coefficient (PCC).
Figure~\ref{fig:xenium-main}b illustrates the corresponding uncertainty
quantification for \textit{TACSTD2}. The confidence maps exhibit
spatially heterogeneous uncertainty: some high-expression regions
remain pronounced at the lower endpoint, whereas other regions show
larger differences between the lower and upper endpoints. Appendix Figure~\ref{fig:xenium-esm-intervals} provides confidence maps
for other genes. Bootstrap-based HIFI similarly yields point-prediction
performance comparable to iStar
(Appendix Figure~\ref{fig:xenium-bootstrap-comparison}) while providing
corresponding confidence interval maps
(Appendix Figure~\ref{fig:xenium-bootstrap-intervals}).

\section{Discussion}

We introduced HIFI, a two-stage framework that separates task adaptation from repeated uncertainty calibration for statistical inference with foundation models. By sharing a Stage~1 task-adapted representation across smaller Stage~2 PEFT fits, HIFI reduces the computational burden of resampling-based inference. Across vision and text tasks, HIFI achieves accurate point estimation and stable near-nominal coverage with substantially less training compute than direct resampling. The Xenium application further demonstrates its utility for uncertainty-aware biological research.

Several limitations remain. First, our experiments focus on LoRA, and extending HIFI to other PEFT parameterizations and resampling methods could broaden its applicability. Second, HIFI relies on PEFT to reduce estimation error, but PEFT effectiveness can be task-dependent, particularly for complex reasoning tasks. Repeated Stage~2 fitting can also remain costly for very large models, motivating extensions to more efficient and reasoning-oriented foundation models.

\section*{Reproducibility Statement}
We provide detailed information to support reproduction of both the theoretical and empirical results. The main text specifies the HIFI framework, inference procedures, and experimental protocols, while the appendix provides the assumptions and complete proofs for the theoretical results. For all experiments, we document the data-generation or preprocessing procedures, model backbones, Stage~1 and Stage~2 data splits, PEFT configurations, optimization settings, resampling schemes, and evaluation protocols. The appendix also reports additional implementation details, hyperparameters, target-wise diagnostics, and supplementary results for the text, vision, and Xenium experiments.

\section*{AI Use Statement}
In this work, we used generative AI tools to generate synthetic data sets, assist with language editing, organization, and formatting of author-written proofs, provide feedback on research methodology or experiments, implement methods, assist with translation, and interpret results.
We did not use generative AI tools to help develop theoretical models or conceptual frameworks, formulate mathematical claims, provide critical ingredients for proving mathematical claims, propose or refine hypotheses, or clean and reformat datasets. Qualitative and thematic data analysis is not applicable to this work. 
We have reviewed all AI-assisted work and take responsibility for the final content of this work, including text, claims, and artifacts produced with the aid of generative AI.
\bibliography{references}
\bibliographystyle{plainnat}

\newpage
\appendix

\section{Theoretical Analysis of HIFI}
\label{app:hifi_theory}

\subsection{Approximation and excess risk}
\label{app:hifi_learning_setup}

Let $\mathcal D_1$ and $\mathcal D_2$ be independent i.i.d. samples
of sizes $n_1$ and $n_2$ from the distribution of $Z=(X,Y)$,
where $r$ is the Stage~2 subsample size.
The target is the link-scale regression function $\theta_0$.
Let $\mathcal X$ be the target domain, with $P_X(\mathcal X)=1$,
and hold the pretrained model fixed. Consider the full adaptation class
\[
    \mathcal F_{\mathbf p}
    =
    \{f(\cdot;w):
      w\in\mathcal W_{\mathbf p}\}.
\]
Here $\mathbf p$ denotes the adaptation specification, which may
encode the number and allocation of trainable parameters, adaptation
ranks, adapted layers, or other architectural complexity measures.
Write $h=(F,t)$ for a Stage~1 output, where
$F$ is the adapted backbone and $t:\mathcal X\to\mathbb R$ is the
corresponding link-scale prediction function. Write
$\widehat h_1=(\widehat F_1,\widehat\theta_1)$ for the fitted
Stage~1 output. For fixed $h$, let
$\mathcal W_{\mathbf p}(h)\subseteq\mathcal W_{\mathbf p}$ denote
the adaptation parameters compatible with the fixed Stage~1 output,
and define $\rho_h=\theta_0-t$. The corresponding Stage~2 residual
class is
\[
    \mathcal R_{\mathbf p}(h)
    =
    \{f(\cdot;w)-t:
      w\in\mathcal W_{\mathbf p}(h)\}.
\]

Throughout, $\|f\|_2^2=\int_{\mathcal X}|f(x)|^2\,dP_X(x)$.
The best approximation errors of the full and Stage~2 classes are
\[
    \mathcal E_{\mathrm{app}}(\mathbf p)
    :=
    \inf_{w\in\mathcal W_{\mathbf p}}
    \|f(\cdot;w)-\theta_0\|_2^2,
    \qquad
    \mathcal E_{\mathrm{app},2}(\mathbf p;h)
    :=
    \inf_{w\in\mathcal W_{\mathbf p}(h)}
    \|f(\cdot;w)-\theta_0\|_2^2.
\]
Equivalently,
$\mathcal E_{\mathrm{app},2}(\mathbf p;h)
=
\inf_{\rho\in\mathcal R_{\mathbf p}(h)}
\|\rho-\rho_h\|_2^2$.
Because $\mathcal W_{\mathbf p}(h)\subseteq\mathcal W_{\mathbf p}$,
the Stage~2 class is a restricted subclass of the full adaptation
class, and hence
\[
    \mathcal E_{\mathrm{app}}(\mathbf p)\le \mathcal E_{\mathrm{app},2}(\mathbf p;h)
    \qquad\text{for every }h.
\]
We therefore define the approximation-capacity loss induced by the
Stage~1 output as
\[
    d_1(h)
    :=
    \mathcal E_{\mathrm{app},2}(\mathbf p;h)-\mathcal E_{\mathrm{app}}(\mathbf p)
    \ge 0.
\]
Thus $d_1(h)$ measures how much approximation capacity is lost by
fixing the Stage~1 output before Stage~2 fitting.

Let $\mathcal E(\mathbf p,m)$ denote a deterministic estimation-error
bound for a learner with adaptation specification $\mathbf p$ fitted
on $m$ observations.

\begin{assumption}
\label{ass:hifi_stage1_learning}
There exists a constant $C_1>0$ such that
\[
    \mathbb E\{d_1(\widehat h_1)\}
    \le
    C_1\mathcal E(\mathbf p,n_1).
\]
\end{assumption}

Here the expectation is over the Stage~1 randomness in
$\widehat h_1$. Assumption~\ref{ass:hifi_stage1_learning}
requires the loss of Stage~2 approximation capacity induced by
the learned Stage~1 output to be controlled by the Stage~1 estimation
error. This is a natural compatibility condition for shared
adaptation: when Stage~1 learns a useful shared backbone, fixing
and reusing it should preserve most of the approximation capacity
available to the full adaptation class.

Let $\widehat\theta^{\mathrm{single}}$ denote a generic single-stage
fit based on a resample of size $c_0r$, where $c_0\ge1$ is fixed, and let
$\widehat\rho(\cdot;h)$ denote a generic Stage~2 residual fit based
on a resample of size $r$. For fixed $h$, let $P_h$ and $E_h$
denote probability and expectation over $\mathcal D_2$ and the
Stage~2 resampling mechanism, with $h$ held fixed. The HIFI
estimator and the corresponding integrated excess risks are
\[
    \widehat\theta^{\mathrm{HIFI}}
    =
    \widehat\theta_1
    +
    \widehat\rho(\cdot;\widehat h_1),
    \qquad
    R_{\mathrm{single}}
    =
    \mathbb E\|\widehat\theta^{\mathrm{single}}-\theta_0\|_2^2,
    \qquad
    R_{\mathrm{HIFI}}
    =
    \mathbb E\|\widehat\theta^{\mathrm{HIFI}}-\theta_0\|_2^2.
\]

\begin{assumption}
\label{ass:hifi_learning_bounds}
There exists a constant $C_{\mathrm{learn}}>0$ such that
\begin{align}
    R_{\mathrm{single}}
    &\le
    C_{\mathrm{learn}}
    \{\mathcal E_{\mathrm{app}}(\mathbf p)
      +\mathcal E(\mathbf p,c_0r)\},
    \label{eq:hifi_single_learning}\\
    E_h\|\widehat\rho(\cdot;h)-\rho_h\|_2^2
    &\le
    C_{\mathrm{learn}}
    \{\mathcal E_{\mathrm{app},2}(\mathbf p;h)+\mathcal E(\mathbf p,r)\},
    \qquad \text{for all } h.
    \label{eq:hifi_conditional_learning}
\end{align}
\end{assumption}

Motivated by oracle inequalities for neural-network regression
\citep[e.g.,][]{schmidt2020nonparametric,meng2025esm}, excess
risk can be bounded by two sources of error: approximation error
and estimation error. We use this structure to compare single-stage
resampling with HIFI. For single-stage resampling, these terms are
$\mathcal E_{\mathrm{app}}(\mathbf p)$ and $\mathcal E(\mathbf p,c_0r)$, respectively; for
HIFI, conditional on the Stage~1 output $h$, they are
$\mathcal E_{\mathrm{app},2}(\mathbf p;h)$ and $\mathcal E(\mathbf p,r)$.

\begin{proposition}
\label{prop:hifi_risk_explicit}
Under Assumptions~\ref{ass:hifi_stage1_learning}
and~\ref{ass:hifi_learning_bounds},
\begin{align}
    R_{\mathrm{single}}
    &\le
    C_{\mathrm{learn}}
    \{\mathcal E_{\mathrm{app}}(\mathbf p)
      +\mathcal E(\mathbf p,c_0r)\},
    \nonumber\\
    R_{\mathrm{HIFI}}
    &\le
    C_{\mathrm{learn}}
    \{\mathcal E_{\mathrm{app}}(\mathbf p)
      +C_1\mathcal E(\mathbf p,n_1)
      +\mathcal E(\mathbf p,r)\}.
    \label{eq:hifi_risk_explicit}
\end{align}
\end{proposition}

\begin{proof}
By the definition of $d_1(h)$,
\[
    \mathcal E_{\mathrm{app},2}(\mathbf p;h)
    =
    \mathcal E_{\mathrm{app}}(\mathbf p)+d_1(h).
\]
Therefore, Assumption~\ref{ass:hifi_stage1_learning} gives
\[
    \mathbb E\{\mathcal E_{\mathrm{app},2}(\mathbf p;\widehat h_1)\}
    \le
    \mathcal E_{\mathrm{app}}(\mathbf p)
    +
    C_1\mathcal E(\mathbf p,n_1).
\]

Since
\[
    \widehat\theta^{\mathrm{HIFI}}-\theta_0
    =
    \widehat\rho(\cdot;\widehat h_1)
    -
    \rho_{\widehat h_1},
\]
the tower property and
Assumption~\ref{ass:hifi_learning_bounds} yield
\begin{align*}
    R_{\mathrm{HIFI}}
    &=
    \mathbb E\left[
        E_{\widehat h_1}
        \left\|
            \widehat\rho(\cdot;\widehat h_1)
            -
            \rho_{\widehat h_1}
        \right\|_2^2
    \right]\\
    &\le
    C_{\mathrm{learn}}
    \left[
        \mathbb E\{\mathcal E_{\mathrm{app},2}(\mathbf p;\widehat h_1)\}
        +
        \mathcal E(\mathbf p,r)
    \right]\\
    &\le
    C_{\mathrm{learn}}
    \{\mathcal E_{\mathrm{app}}(\mathbf p)
      +C_1\mathcal E(\mathbf p,n_1)
      +\mathcal E(\mathbf p,r)\}.
\end{align*}
The single-stage bound follows directly from
\eqref{eq:hifi_single_learning}.
\end{proof}

The bounds reveal the key distinction between single-stage
resampling and HIFI. Single-stage resampling fits the full adaptation
class $\mathcal F_{\mathbf p}$ on each resample, leading to the
estimation term $\mathcal E(\mathbf p,c_0r)$. HIFI instead pays the
Stage~1 estimation cost $\mathcal E(\mathbf p,n_1)$ once and,
conditional on the Stage~1 output, resampling fits only the restricted
Stage~2 class, leading to $\mathcal E(\mathbf p,r)$. Both
procedures retain the same full-class approximation term
$\mathcal E_{\mathrm{app}}(\mathbf p)$. Therefore, when the two procedures attain risk
bounds of the same order, across $B$ resampling fits HIFI incurs
the Stage~1 estimation term only once and the Stage~2 estimation term
$B$ times, rather than refitting the full adaptation class
$\mathcal F_{\mathbf p}$ $B$ times as in single-stage resampling.

\subsection{Inferential validity}
\label{app:hifi_pointwise_control}

Valid inference requires, in addition to the excess-risk bounds,
the regularity conditions of the chosen resampling method.

\paragraph{HIFI--ESM validity}
Let $\mathcal H_{n_2}$ be deterministic sets of Stage~1 outputs
$h$ for which the conditional Stage~2 DNN--ESM problem for
the residual target $\rho_h$ satisfies the conditions in
Corollary~3.6 and Theorems~3.7--3.8 of \citet{meng2025esm}.
We further assume that there exist common deterministic target sets
$\mathcal T_{n_2}\subseteq\mathcal X$ satisfying
\[
    P_X(\mathcal T_{n_2})\ge1-\delta_{n_2},
    \qquad
    \delta_{n_2}>0,
    \qquad
    \delta_{n_2}\to0,
\]
on which the resulting conditional normal approximation centered
at the true residual target $\rho_h$ and variance consistency
hold uniformly over $h\in\mathcal H_{n_2}$.

To account for the randomness of the Stage~1 fit, assume that
\[
    P(\widehat h_1\in\mathcal H_{n_2})\to1.
\]

Under this uniform conditional formulation, Slutsky's theorem
gives, for every deterministic sequence
$x_{n_2}\in\mathcal T_{n_2}$,
\begin{equation}
    \sup_{h\in\mathcal H_{n_2}}
    \left|
        P_h\left\{
            \theta_0(x_{n_2})
            \in
            C_{\alpha}^{\theta,\mathrm{ESM}}(x_{n_2};h)
        \right\}
        -(1-\alpha)
    \right|
    \to0.
    \label{eq:hifi_uniform_conditional_validity}
\end{equation}
On the event $\widehat h_1\in\mathcal H_{n_2}$, the conditional
coverage error at $\widehat h_1$ is bounded by the supremum in
\eqref{eq:hifi_uniform_conditional_validity}. Hence,
\[
    P_{\widehat h_1}
    \left\{
        \theta_0(x_{n_2})
        \in
        C_{\alpha}^{\theta,\mathrm{ESM}}
        (x_{n_2};\widehat h_1)
    \right\}
    \to_p
    1-\alpha.
\]
Unconditional validity follows from the following lemma.

\begin{lemma}
\label{lem:hifi_unconditional_validity}
Suppose that, for a deterministic target sequence $x_{n_2}$ and
a sequence of confidence intervals
$C_\alpha(x_{n_2};\widehat h_1)$,
\[
    P_{\widehat h_1}
    \left\{
        \theta_0(x_{n_2})
        \in
        C_\alpha(x_{n_2};\widehat h_1)
    \right\}
    \to_p
    1-\alpha.
\]
Then
\[
    P
    \left\{
        \theta_0(x_{n_2})
        \in
        C_\alpha(x_{n_2};\widehat h_1)
    \right\}
    \to
    1-\alpha.
\]
\end{lemma}

\begin{proof}
Let
\[
    q_{n_2}(\widehat h_1)
    :=
    P_{\widehat h_1}
    \left\{
        \theta_0(x_{n_2})
        \in
        C_\alpha(x_{n_2};\widehat h_1)
    \right\}.
\]
By assumption,
$q_{n_2}(\widehat h_1)\to_p1-\alpha$.
Since $0\le q_{n_2}(\widehat h_1)\le1$, convergence in probability to
the constant $1-\alpha$ implies convergence in $L_1$. Hence,
\[
    \mathbb E\{q_{n_2}(\widehat h_1)\}
    \to
    1-\alpha.
\]
By the tower property,
\[
    \mathbb E\{q_{n_2}(\widehat h_1)\}
    =
    P
    \left\{
        \theta_0(x_{n_2})
        \in
        C_\alpha(x_{n_2};\widehat h_1)
    \right\},
\]
which proves the result.
\end{proof}

Applying Lemma~\ref{lem:hifi_unconditional_validity} gives the
HIFI--ESM coverage statement in the main text.

\paragraph{HIFI--Bootstrap validity}
For bootstrap inference, assume that, for each deterministic
admissible Stage~1 output $h$, the conditional sampling law
of the normalized Stage~2 estimation error centered at
$\rho_h(x_{n_2})$ and the bootstrap law of the
correspondingly normalized error centered at the original
Stage~2 estimate converge in probability to the same
distribution, with a continuous, strictly increasing cdf
and symmetry about zero. Assume that the number of conditionally
independent bootstrap replicates tends to infinity.
The bootstrap quantile argument of \citet[Section~4, Lemma~4.2]{bucher2019note},
together with the symmetry assumption, then yields asymptotically
correct conditional coverage for percentile intervals.
Under the analogous uniform conditional coverage requirement over
$\mathcal H_{n_2}$ and $\mathcal T_{n_2}$, evaluating at
$\widehat h_1$ and applying
Lemma~\ref{lem:hifi_unconditional_validity} gives unconditional
validity.

\section{Implementation Details}
\label{app:inference-implementation}

The implementation of HIFI with ESM is described in
Section~\ref{sec:instantiation}. Here we provide the corresponding bootstrap construction and implementation details for the external comparators.

\subsection{Bootstrap}
\label{app:bootstrap_implementation}

We use the LoRA construction in Section~\ref{sec:instantiation},
holding the Stage~1 fit fixed. For Stage~2, we independently draw
$B$ bootstrap samples, each containing $n_2$ observations sampled
with replacement from $\mathcal D_2$. The corresponding weights satisfy
\[
    (W_{b1},\ldots,W_{bn_2})
    \overset{\mathrm{iid}}{\sim}
    \operatorname{Multinomial}
    \left(n_2;\frac{1}{n_2},\ldots,\frac{1}{n_2}\right),
    \qquad b=1,\ldots,B,
\]
where $W_{bi}$ counts the occurrences of observation $i$ in sample $b$.
Each fit optimizes new Stage~2 LoRA factors and a prediction head
according to \eqref{eq:stage2_objective}, yielding $\widehat\rho_b$.

We average the $B$ residual estimators,
\[
    \widehat\rho_{\mathrm{Boot}}(x)
    =
    \frac{1}{B}\sum_{b=1}^B\widehat\rho_b(x),
    \qquad
    \widehat\theta_{\mathrm{Boot}}(x)
    =
    \widehat\theta_1(x)
    +
    \widehat\rho_{\mathrm{Boot}}(x).
\]

Let $\widehat Q_p^\rho(x)$ denote the empirical $p$-quantile of
$\{\widehat\rho_b(x)\}_{b=1}^B$. The pointwise percentile interval
for $\theta_0(x)$ is
\[
    C_{\alpha}^{\theta,\mathrm{Boot}}(x)
    =
    \widehat\theta_1(x)
    +
    \left[
        \widehat Q_{\alpha/2}^\rho(x),
        \widehat Q_{1-\alpha/2}^\rho(x)
    \right].
\]
The point estimate and interval are transformed to the response
scale through $g^{-1}$.

\subsection{External Comparators}

\paragraph{Bayesian-PEFT} We adapt the Bayesian neural-network framework of \citet{sun2021sparse} to parameter-efficient fine-tuning. The pretrained backbone is held fixed, and approximate posterior sampling is performed over the LoRA parameters and prediction head using stochastic-gradient Hamiltonian Monte Carlo \citep{chen2014stochastic}. The model is fitted
directly to the observed responses. Text experiments use a Bernoulli likelihood, whereas vision experiments use a Gaussian likelihood with an estimated global noise scale. For each target, the point estimate is the mean of the retained conditional-mean predictions, and empirical equal-tailed quantiles form approximate posterior credible intervals.
Model-specific sampling and prior settings are reported in
Tables~\ref{tab:vision_external_settings}
and~\ref{tab:text_external_settings}.

\paragraph{HulC}

We implement HulC \citep{kuchibhotla2024hulc} as a direct comparator.
The labeled data are partitioned into disjoint subsets, and an
independent LoRA estimator is fitted to each subset from the pretrained
backbone. For a fixed target $x$, let
$\widehat\theta^{\mathrm{HulC}}_b(x)$, $b=1,\ldots,B_\alpha$,
denote the split-specific link-scale estimates. The link-scale
interval is
\[
C_\alpha^{\theta,\mathrm{HulC}}(x)
=
\left[
\min_{1\le b\le B_\alpha}\widehat\theta^{\mathrm{HulC}}_b(x),
\max_{1\le b\le B_\alpha}\widehat\theta^{\mathrm{HulC}}_b(x)
\right],
\]
and the response-scale interval is obtained through $g^{-1}$.
We use the median of the split-specific estimates as the point estimate
and the $\Delta=0$ construction, with the number of disjoint fits chosen
for the desired nominal confidence level.

\section{Synthetic Experiment Details}

\subsection{Synthetic Vision Data}
\label{app:vision-dgp}

\begin{figure}[t]
    \centering
    \includegraphics[width=\textwidth]
    {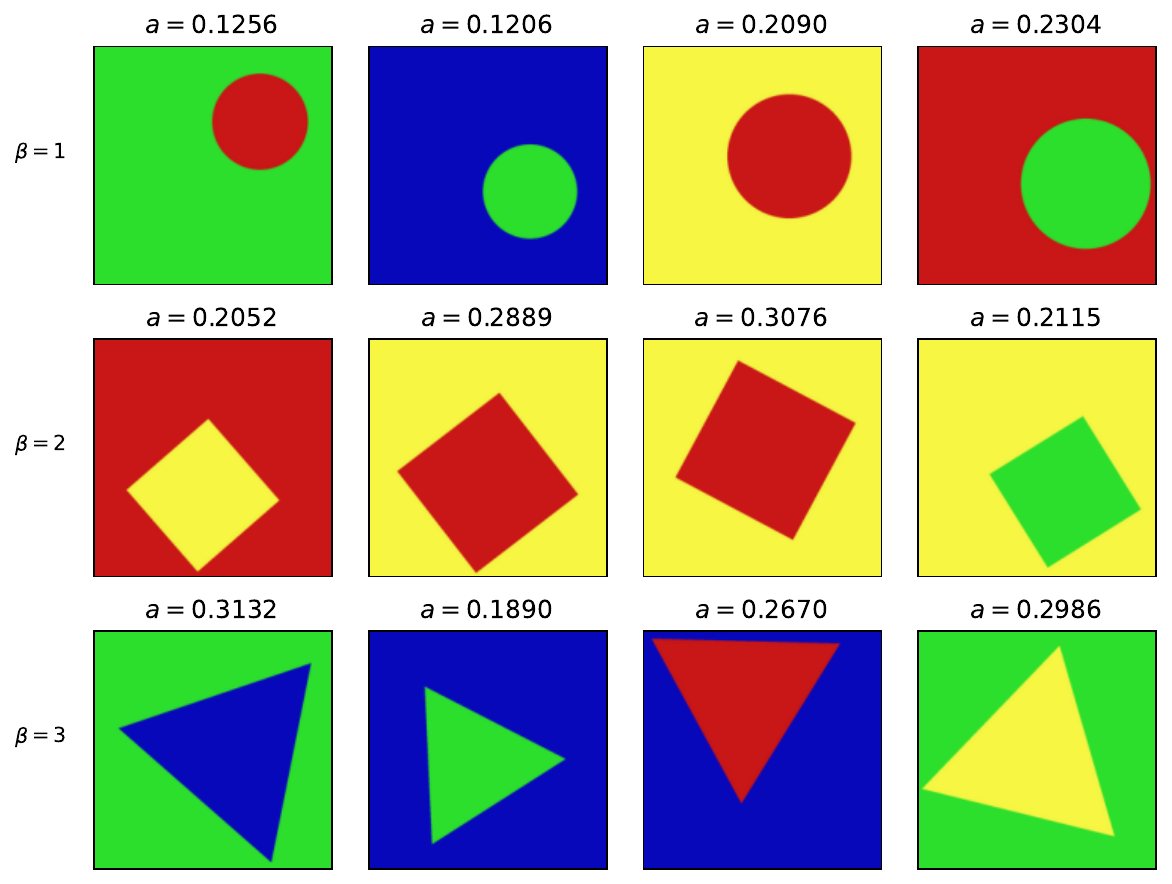}
    \caption{
        Examples of the synthetic vision data.
        Rows correspond to circles, squares, and equilateral
        triangles, respectively.
        Foreground and background colors, object positions,
        and polygon orientations vary across images.
    }
    \label{fig:vit-simulation-examples}
\end{figure}

We provide additional data-generation details for the synthetic
image-regression task in Section~\ref{sec:vision-task}.
Each object is defined on a unit-square canvas, and $a$ denotes
its geometric area relative to the canvas. Triangles are equilateral,
while squares and triangles are rotated by an angle sampled uniformly
from $[0,2\pi)$. Conditional on shape, area, and orientation, the
object center is sampled uniformly over locations that keep the entire
object inside the canvas. Foreground and background colors are sampled
uniformly from the 12 ordered distinct pairs formed by red $(255,0,0)$,
green $(0,255,0)$, yellow $(255,255,0)$, and blue $(0,0,255)$,
independently of shape and area. For each fixed target image $x_j$,
evaluation uses the conditional-mean target $\beta(S_j)a_j$.
A panel of 500 target images is generated once and held fixed across all 100 replications. Representative examples are shown in
Figure~\ref{fig:vit-simulation-examples}.

\subsection{Bilingual Text Data Generation}
\label{app:text-dgp}

We provide the data generation details for the bilingual simulation described in Section~\ref{sec:text-task}.
\paragraph{Probability-stratified sampling.}
The binary outcome probability is determined by the arrival lead time
through
\[
    \mu(T)
    =
    \operatorname{logit}^{-1}\{2.8(T-1.05)\},
    \qquad T\in[0,3.5].
\]
Thus, the attainable probability range is
\[
    [\mu_{\min},\mu_{\max}]
    =
    [\mu(0),\mu(3.5)]
    =
    [\operatorname{logit}^{-1}(-2.94),
      \operatorname{logit}^{-1}(6.86)].
\]

To place equal sampling probability across regions of the success-probability range
and avoid concentrating observations in only part of this range,
we partition $[\mu_{\min},\mu_{\max}]$ into 20 equal-width probability
strata. For each observation, we independently draw a stratum index
\[
    J \sim \operatorname{Uniform}\{1,\ldots,20\}.
\]
Let
\[
    a_j
    =
    \mu_{\min}
    +
    \frac{j}{20}(\mu_{\max}-\mu_{\min}),
    \qquad j=0,\ldots,20,
\]
denote the stratum boundaries, so that stratum $j$ corresponds to
$(a_{j-1},a_j]$. Conditional on $J=j$, we sample
\[
    p^\star
    \sim
    \operatorname{Uniform}(a_{j-1},a_j),
\]
and invert the probability model to obtain the corresponding arrival
lead time, rounded to the nearest minute:
\[
    T
    =
    \frac{
        \operatorname{round}
        \left\{
            60\left[
                1.05+\frac{\operatorname{logit}(p^\star)}{2.8}
            \right]
        \right\}
    }{60}.
\]
\paragraph{Text construction.}

For each observation, independently of $T$, the language is drawn as
English or Chinese with equal probability. Conditional on $T$ and the
sampled language, we generate a natural-language description of when
the passenger arrives relative to the scheduled departure time.
Additional textual variation is introduced through auxiliary flight
information, such as the airline, flight number, gate, terminal, seat,
or boarding group, with probability $0.50$, and through
response-irrelevant contextual information with probability $0.20$.
These components are generated independently of the arrival lead time.
The sampled lead time is expressed in the resulting text input $X$;
by construction, $T(X)=T$.

\paragraph{Response generation.}
For the resulting text $X$, we set
\[
    \mu_0(X)
    =
    \operatorname{logit}^{-1}
    \{2.8[T(X)-1.05]\},
\]
and independently draw
\[
    Y\mid X
    \sim
    \operatorname{Bernoulli}\{\mu_0(X)\},
\]
where $Y=1$ indicates that the passenger catches the flight and
$Y=0$ otherwise. Table~\ref{tab:text-dgp-examples} provides
representative realizations from the bilingual DGP, showing
the generated text $X$, the encoded arrival lead time $T(X)$, the
corresponding conditional mean $\mu_0(X)$, and the sampled outcome $Y$. The model is trained only on the observed pairs $(X,Y)$ and is not given $T(X)$ or $\mu_0(X)$.

\begin{CJK}{UTF8}{gbsn}
\begin{table}[t]
\centering
\footnotesize
\caption{Illustrative observations from the bilingual text DGP
at representative arrival lead times.  English translations of the Chinese inputs are included only for readability.}
\label{tab:text-dgp-examples}
\setlength{\tabcolsep}{4pt}
\renewcommand{\arraystretch}{1.18}
\begin{tabularx}{\textwidth}{@{} l >{\raggedright\arraybackslash}X c c c @{}}
\toprule
Language & Model input $X$ & $T(X)$ (min) & $\mu_0(X)$ & $Y$ \\
\midrule
English & Kate walked into the terminal 15 minutes before the scheduled departure. & 15 & 0.0962 & 0 \\
English & Zack arrived at the airport 49 minutes before his flight. & 49 & 0.3422 & 1 \\
English & His flight DL8517 was assigned to gate B17. Henry reached the gate 1 hour and 16 minutes ahead of takeoff. & 76 & 0.6472 & 0 \\
English & American listed the flight as AA2789 from Terminal A. Frank reached the airport with 1 hour and 50 minutes left until the scheduled flight. & 110 & 0.8996 & 1 \\
\midrule
Chinese & 航班预定起飞前16分钟，周强已经抵达机场。\par {\scriptsize\itshape English translation: Zhou Qiang had already arrived at the airport 16 minutes before the scheduled departure.} & 16 & 0.1004 & 0 \\
Chinese & 航站楼里人特别多。 她的航班是长荣BR6967。 距起飞还剩50分钟时，林雪走进了候机厅。\par {\scriptsize\itshape English translation: The terminal was especially crowded. Her flight was EVA Air BR6967. With 50 minutes remaining before takeoff, Lin Xue entered the departure lounge.} & 50 & 0.3528 & 0 \\
Chinese & 他的航班是东航MU5511。 航班预定起飞前1小时16分钟，郑磊已经抵达机场。\par {\scriptsize\itshape English translation: His flight was China Eastern MU5511. Zheng Lei had already arrived at the airport 1 hour and 16 minutes before the scheduled departure.} & 76 & 0.6472 & 1 \\
Chinese & 她的航班从一号航站楼出发。 查完行程后，吴丽在飞机起飞前1小时50分钟到达了机场。\par {\scriptsize\itshape English translation: Her flight departed from Terminal 1. After checking the itinerary, Wu Li arrived at the airport 1 hour and 50 minutes before takeoff.} & 110 & 0.8996 & 1 \\
\bottomrule
\end{tabularx}
\end{table}
\end{CJK}

\section{Model Architecture and Training Settings}
\label{app:experiment-details}

Across both modalities, the learned Stage~1 LoRA updates are merged into the backbone and frozen before Stage~2, which uses fresh LoRA adapters and a residual prediction head with a zero-initialized output layer. The remaining architecture, adaptation, optimization, and inference settings are reported below.

\subsection{Vision Experiment}
\label{app:vision-experiment-details}

HIFI settings are summarized in Table~\ref{tab:vision_model_settings}, and settings for HulC and
Bayesian-PEFT are reported in Table~\ref{tab:vision_external_settings}.

\subsection{Text Experiment}
\label{app:text-experiment-details}

HIFI settings are summarized in Table~\ref{tab:text_model_settings}, and settings for HulC and
Bayesian-PEFT are reported in Table~\ref{tab:text_external_settings}.

\subsection{Ablation Study 1}

We use Qwen3-Embedding-8B with the architecture, LoRA
configuration, and optimization hyperparameters reported in
Table~\ref{tab:text_model_settings}. We vary the total Stage~1 data budget over
$\{2{,}000,4{,}500,8{,}500\}$, maintaining the same
training-to-validation ratio as in the main experiment, while keeping the Stage~2
architecture and data budget fixed. The single-stage reference uses the same
member architecture and optimization settings as HIFI
Stage~2, but fits the response directly from the pretrained
backbone with a standard randomly initialized prediction
head.

\subsection{Ablation Study 2}

The HIFI configurations follow Tables~\ref{tab:vision_model_settings}
and~\ref{tab:text_model_settings} for the corresponding
backbones. Single-stage learners use the corresponding
HIFI Stage~2 architecture, LoRA configuration, and
optimization hyperparameters. Each Single-stage learner
uses a training pool of 9,000 observations and a validation
pool of 700 observations. For single-stage ESM, we draw 500 subsamples of size
3,621 without replacement from the
9,000 training observations. For single-stage bootstrap,
we draw 400 resamples of size 9000
with replacement from the same training pool. HIFI allocates these observations
to Stage~1 (8,000 training and 500 validation) and
Stage~2 (1,000 training and 200 validation).

\begin{table}[t]
\centering
\caption{Architecture, adaptation, and training settings for the vision experiments.}
\label{tab:vision_model_settings}
\small
\setlength{\tabcolsep}{5pt}
\renewcommand{\arraystretch}{1.08}
\begin{tabularx}{\textwidth}{@{}m{0.30\textwidth} *{3}{>{\centering\arraybackslash}X}@{}}
\toprule
Setting & ViT-B/32 & DINOv2-B/14 & Swin-B \\
\midrule

Pooling
& CLS + mean patch (concatenated)
& CLS + mean patch (concatenated)
& Global average \\

Prediction head
& \multicolumn{3}{c}{Two-layer MLP; 64 hidden units; GELU} \\

Training loss
& \multicolumn{3}{c}{Mean squared error} \\

\addlinespace[4pt]
\multicolumn{4}{@{}l}{\textit{Stage 1}} \\
\cmidrule(l){2-4}

Training / validation sample size
& \multicolumn{3}{c}{8,000 / 500} \\

Blocks with LoRA
& All 12
& Last 4
& All 24 \\

LoRA projections / dropout
& \multicolumn{3}{c}{Q, K, V, and O; dropout 0.05} \\

LoRA rank / scaling
& Rank 12; scaling 24
& Rank 4;  scaling 8
& Rank 12; scaling 24 \\

Learning rate
& $5\times10^{-5}$
& $4\times10^{-4}$
& $5\times10^{-5}$ \\

Training settings
& \multicolumn{3}{c}{
AdamW; batch size 16; weight decay $10^{-4}$; at most 80 epochs} \\

\addlinespace[4pt]
\multicolumn{4}{@{}l}{\textit{Stage 2}} \\
\cmidrule(l){2-4}

Training / validation sample size
& \multicolumn{3}{c}{1,000 / 200} \\

Blocks with LoRA
& Last 6
& Last 2
& Final stage (2 blocks) \\

LoRA configuration
& \multicolumn{3}{c}{
Q and V projections; rank 8; scaling 16; dropout 0.05} \\

Learning rate
& $10^{-4}$
& $5\times10^{-5}$
& $10^{-4}$ \\

Training settings
& \multicolumn{3}{c}{
AdamW; batch size 8 per member; weight decay $10^{-4}$} \\

Maximum epochs
& 40
& 25
& 25 \\

\addlinespace[4pt]
\multicolumn{4}{@{}l}{\textit{Stage-2 inference}} \\
\cmidrule(l){2-4}

ESM subsamples / size
& \multicolumn{3}{c}{500 / 501; without replacement} \\

Bootstrap resamples / size
& \multicolumn{3}{c}{400 / 1,000; with replacement} \\

\bottomrule
\end{tabularx}
\end{table}

\begin{table}[t]
\centering
\caption{Settings for the external vision comparators.}
\label{tab:vision_external_settings}
\small
\setlength{\tabcolsep}{3pt}
\renewcommand{\arraystretch}{1.08}
\begin{tabularx}{\linewidth}{@{}p{0.30\linewidth}*{3}{>{\centering\arraybackslash}X}@{}}
\toprule
Setting & ViT-B/32 & DINOv2-B/14 & Swin-B \\
\midrule
\multicolumn{4}{@{}l}{\textit{HulC}} \\
Training / validation size
& \multicolumn{3}{c}{9,000 / 700} \\
Adapted blocks
& Last 6 & Last 2 & Final stage (2 blocks) \\
LoRA configuration
& \multicolumn{3}{c}{Q and V; rank 8; scaling 16; dropout 0.05} \\
Learning rate
& $10^{-4}$ & $5\times10^{-5}$ & $10^{-4}$ \\
Training
& \multicolumn{3}{c}{AdamW; batch size 8; weight decay $10^{-4}$} \\
Maximum epochs
& 40 & 25 & 25 \\
Disjoint fits (90\% / 95\%)
& \multicolumn{3}{c}{5 / 6} \\
\midrule
\multicolumn{4}{@{}l}{\textit{Bayesian-PEFT}} \\
Training size
& \multicolumn{3}{c}{9,700} \\
Adapted blocks
& All 12 & All 12 & All 24 \\
LoRA configuration
& \multicolumn{3}{c}{Q, K, V, and O; rank 12; scaling 24} \\
SGHMC learning rate
& $10^{-4}$ & $5\times10^{-5}$ & $5\times10^{-5}$ \\
SGHMC sampling
& \multicolumn{3}{c}{Batch size 8; momentum 0.9} \\
Posterior samples
& \multicolumn{3}{c}{600 burn-in updates; thinning 30; 300 draws} \\
Weight prior
& \multicolumn{3}{c}{Zero-mean Gaussian spike-and-slab mixture} \\
Spike / slab variance
& \multicolumn{3}{c}{$10^{-6}$ / $10^{-2}$} \\
Slab probability
& \multicolumn{3}{c}{$10^{-7}$} \\
Likelihood
& \multicolumn{3}{c}{Gaussian; observation SD $\sigma$ estimated from data} \\
Observation-SD prior
& \multicolumn{3}{c}{$\sigma\sim\mathrm{HalfNormal}(1)$} \\
\bottomrule
\end{tabularx}
\vspace{2pt}
\begin{minipage}{\linewidth}
\footnotesize
\textit{Notes.}
Both comparators use the backbone-specific pooling and prediction head
reported in Table~\ref{tab:vision_model_settings}.
\end{minipage}
\end{table}

\begin{table}[t]
\centering
\begin{threeparttable}
\caption{Architecture, adaptation, and training settings for the text experiments.}
\label{tab:text_model_settings}
\small
\setlength{\tabcolsep}{5pt}
\renewcommand{\arraystretch}{1.08}
\renewcommand{\tabularxcolumn}[1]{m{#1}}
\begin{tabularx}{\textwidth}{@{}m{0.30\textwidth} *{3}{>{\centering\arraybackslash}X}@{}}
\toprule
Setting & Qwen3-Embedding-8B & XLM-R XXL & NV-Embed-v2 \\
\midrule

Pooling
& Last token
& CLS token
& Native latent attention \\

Maximum sequence length
& 96 tokens
& 96 tokens
& 112 tokens \\

Input prompt
& \multicolumn{3}{>{\raggedright\arraybackslash}p{\dimexpr0.70\textwidth-2\tabcolsep\relax}@{}}{%
\footnotesize\ttfamily
Instruct: Given a passenger flight-status description, estimate the
probability that the passenger makes the flight.\newline
Query: <text>} \\

Prediction head
& \multicolumn{3}{c}{Two-layer MLP; 128 hidden units; GELU} \\

Training loss
& \multicolumn{3}{c}{Binary cross-entropy with logits} \\

LoRA configuration
& \multicolumn{3}{c}{Q and V projections; rank 16; scaling 32; dropout 0.05} \\

\addlinespace[4pt]
\multicolumn{4}{@{}l}{\textit{Stage 1}} \\
\cmidrule(l){2-4}

Training / validation sample size
& \multicolumn{3}{c}{8,000 / 500} \\

Blocks with LoRA
& All 36
& All 48
& All 32 \\

Learning rate
& $7.5\times10^{-5}$
& $5\times10^{-5}$
& $3\times10^{-5}$ \\

Training settings
& \multicolumn{3}{c}{
AdamW; batch size 8; weight decay $10^{-4}$; at most 20 epochs} \\

\addlinespace[4pt]
\multicolumn{4}{@{}l}{\textit{Stage 2}} \\
\cmidrule(l){2-4}

Training / validation sample size
& \multicolumn{3}{c}{1,000 / 200} \\

Blocks with LoRA
& Last 6
& Last 4
& Last 4 \\

Learning rate
& $4\times10^{-5}$
& $10^{-4}$
& $3\times10^{-4}$ \\

Training settings
& \multicolumn{3}{c}{
AdamW; batch size 8 per member; weight decay 0; at most 25 epochs} \\

\addlinespace[4pt]
\multicolumn{4}{@{}l}{\textit{Stage-2 inference}} \\
\cmidrule(l){2-4}

ESM subsamples / size
& \multicolumn{3}{c}{500 / 501; without replacement} \\

Bootstrap resamples / size
& \multicolumn{3}{c}{400 / 1,000; with replacement} \\

\bottomrule
\end{tabularx}
\end{threeparttable}
\end{table}

\begin{table}[t]
\centering
\caption{Settings for the external text comparators.}
\label{tab:text_external_settings}
\small
\setlength{\tabcolsep}{3pt}
\renewcommand{\arraystretch}{1.08}
\begin{tabularx}{\linewidth}{@{}p{0.29\linewidth}*{3}{>{\centering\arraybackslash}X}@{}}
\toprule
Setting & Qwen3-Embedding-8B & XLM-R XXL & NV-Embed-v2 \\
\midrule
\multicolumn{4}{@{}l}{\textit{HulC}} \\
Training / validation size
& \multicolumn{3}{c}{9,000 / 700} \\
Adapted blocks
& Last 6 & Last 4 & Last 4 \\
LoRA configuration
& \multicolumn{3}{c}{Q and V; rank 16; scaling 32; dropout 0.05} \\
Learning rate
& $4\times10^{-5}$ & $2\times10^{-4}$ & $3\times10^{-4}$ \\
Training
& \multicolumn{3}{c}{AdamW; batch size 8; weight decay 0} \\
Maximum epochs
& \multicolumn{3}{c}{25} \\
Disjoint fits (90\% / 95\%)
& \multicolumn{3}{c}{5 / 6} \\
\midrule
\multicolumn{4}{@{}l}{\textit{Bayesian-PEFT}} \\
Training size
& \multicolumn{3}{c}{9,700} \\
Adapted blocks
& All 36 & Last 12 & Last 12 \\
LoRA configuration
& \multicolumn{3}{c}{Q and V; rank 16; scaling 32} \\
Likelihood
& \multicolumn{3}{c}{Bernoulli with a logit link} \\
SGHMC learning rate
& $5\times10^{-5}$ & $5\times10^{-3}$ & $5\times10^{-4}$ \\
SGHMC sampling
& \multicolumn{3}{c}{Batch size 8; momentum 0.9; temperature 1} \\
Burn-in updates
& 600 & 1,200 & 600 \\
Posterior samples
& \multicolumn{3}{c}{300 draws; thinning interval 30} \\
Parameter prior
& Gaussian spike-and-slab mixture
& $\mathcal{N}(0,1)$
& 
$\mathcal{N}(0,10^{-2})$: LoRA and head weights;
$\mathcal{N}(0,1)$: head biases \\
\bottomrule
\end{tabularx}
\vspace{2pt}
\begin{minipage}{\linewidth}
\footnotesize
\textit{Notes.}
Both comparators use the pooling, input prompt, sequence-length limit, and prediction-head architecture in Table~\ref{tab:text_model_settings}.
\end{minipage}
\end{table}

\section{Additional Synthetic Experiment Results}

\subsection{Target-level Results for the Vision Task}
\label{app:vision-target-diagnostics}

\begin{figure}[t]
\centering
\includegraphics[width=\textwidth]
{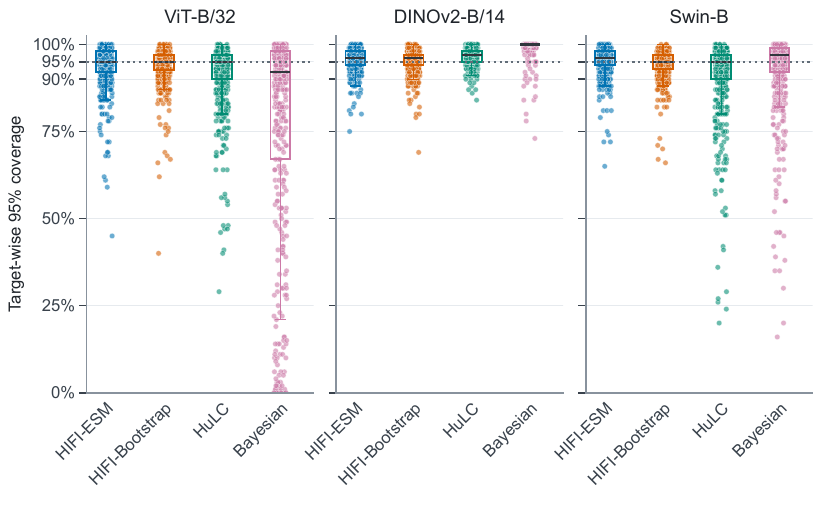}
\caption{
Target-wise empirical coverage of nominal 95\% pointwise confidence intervals for the vision regression task. Each point represents one fixed evaluation target and gives the proportion of replications in which the corresponding interval contains $\mu_0(x)$. The dashed horizontal line marks the nominal 95\% coverage level.
}
\label{fig:vision-target-coverage}
\end{figure}

\begin{figure}[t]
\centering
\includegraphics[width=\textwidth]
{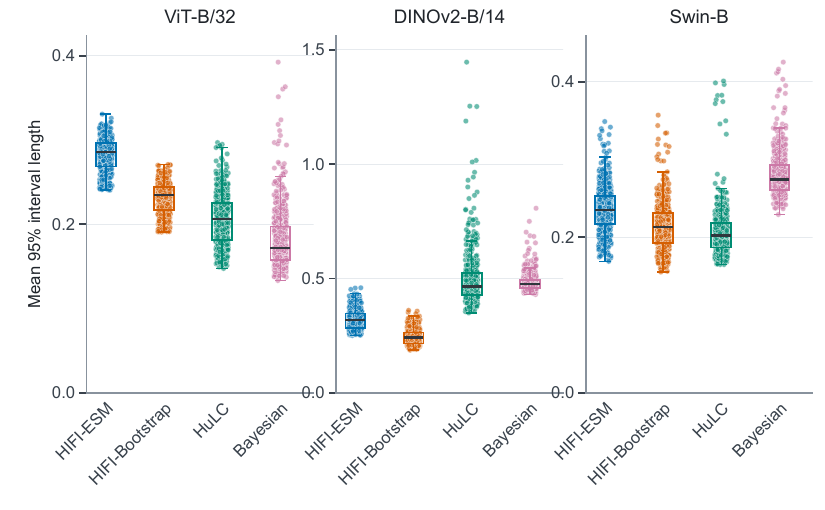}
\caption{
Target-wise mean lengths of nominal 95\% pointwise confidence intervals for the vision regression task. Each point represents one evaluation target and reports the confidence-interval length averaged across replications. Boxplots summarize the distribution of mean interval lengths across targets.
}
\label{fig:vision-target-width}
\end{figure}

Figure~\ref{fig:vision-target-coverage} reports empirical 95\%
coverage across individual evaluation targets, while Figure~\ref{fig:vision-target-width} summarizes the corresponding distribution of mean 95\% interval lengths. HIFI-ESM and HIFI-Bootstrap exhibit stable target-wise calibration across all three vision backbones, with coverage concentrated near the nominal level for most targets and only a limited number showing noticeable under-coverage. In contrast, Bayesian-PEFT exhibits pronounced under-coverage for many targets under ViT-B/32, while HulC shows heavier lower tails in several settings. In terms of interval length, HIFI-Bootstrap generally produces shorter intervals than HIFI-ESM, while both methods maintain relatively stable interval lengths across targets. These results show that HIFI provides well-calibrated uncertainty quantification across evaluation targets
while maintaining competitive interval efficiency.

\begin{figure}[t]
\centering
\includegraphics[width=\textwidth]
{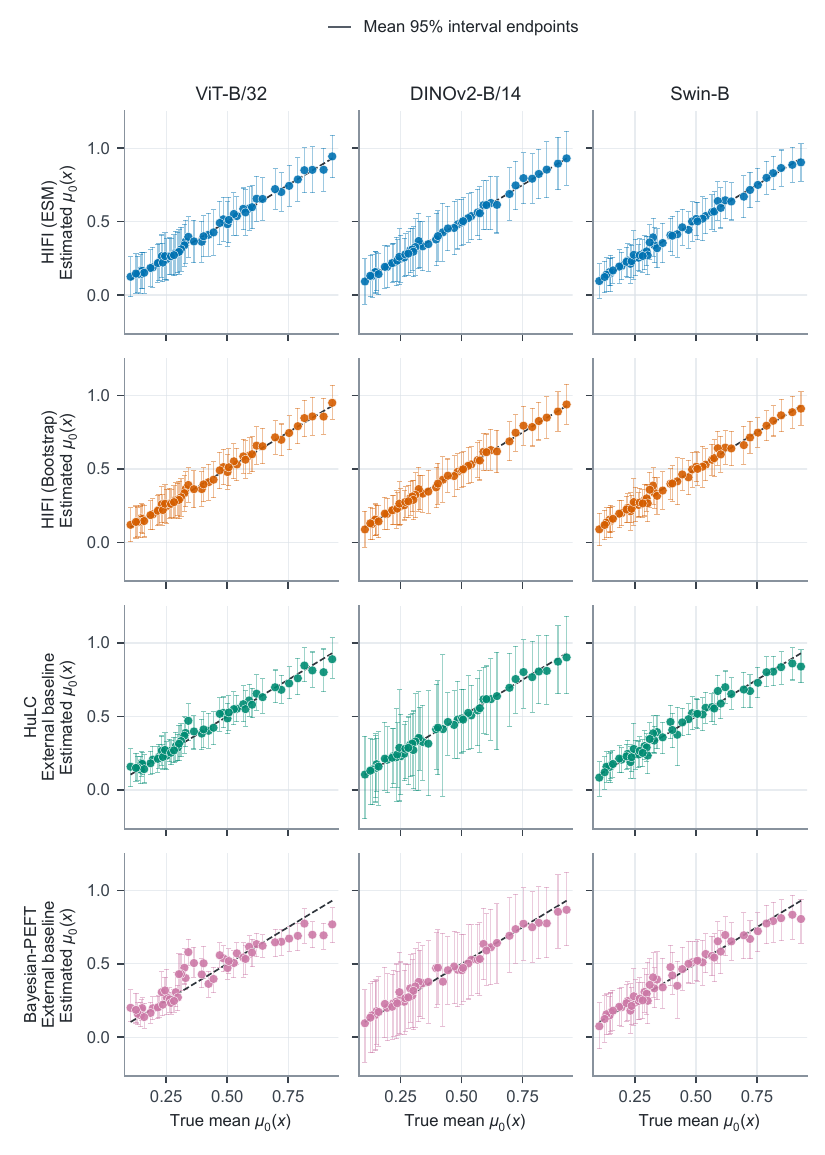}
\caption{
For visualization, 50 targets are selected from the full target panel by stratified random sampling across shape classes and true-mean bins. 
Columns correspond to the three vision backbones, and rows correspond to HIFI-ESM, HIFI-Bootstrap, HulC, and Bayesian-PEFT. Points show the mean estimated conditional mean for each target, and
vertical segments show the mean lower and upper endpoints of the corresponding 95\% intervals.
The dashed diagonal line represents perfect agreement with the true
conditional mean $\mu_0(x)$.
}
\label{fig:vision-target-predictions}
\end{figure}

For visualization, 50 targets are selected from the full target panel by stratified random sampling across shape classes and true-mean bins. Figure~\ref{fig:vision-target-predictions} shows the corresponding mean point estimates and mean endpoints of the 95\% confidence intervals across replications. Across all three backbones, the HIFI estimators closely track the identity line over the range of true conditional means. The external baselines exhibit noticeable target-specific bias, with mean estimates for some targets deviating substantially from the identity line.

\subsection{Target-level Results for the Text Task}
\label{app:text-target-diagnostics}

\begin{figure}[t]
\centering
\includegraphics[width=\textwidth]
{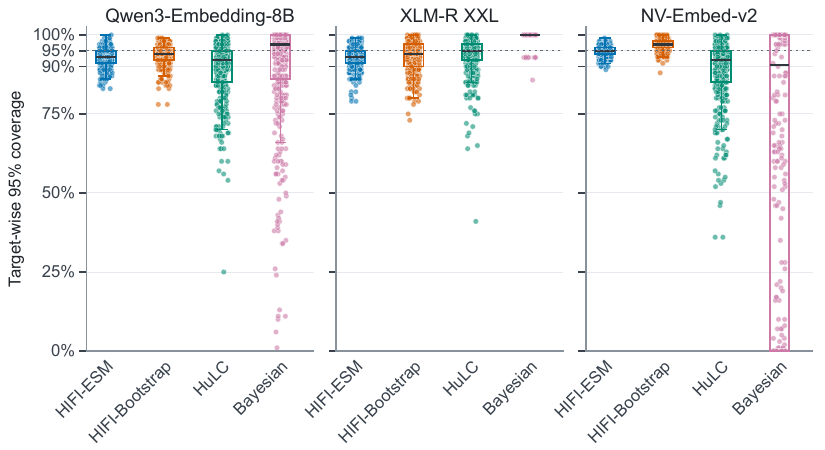}
\caption{
Target-wise empirical coverage of nominal 95\% pointwise confidence
intervals for the text probability-prediction task. Panels correspond
to the three language backbones, and within each panel results are
shown for HIFI-ESM, HIFI-Bootstrap, HulC, and Bayesian-PEFT. Each
point represents one fixed evaluation target and gives the proportion
of replications in which the corresponding interval
contains the true conditional probability $\mu_0(x)$. Boxplots
summarize the distribution across targets, and the dashed horizontal
line marks the nominal 95\% coverage level.
}
\label{fig:text-target-coverage}
\end{figure}

\begin{figure}[t]
\centering
\includegraphics[width=\textwidth]
{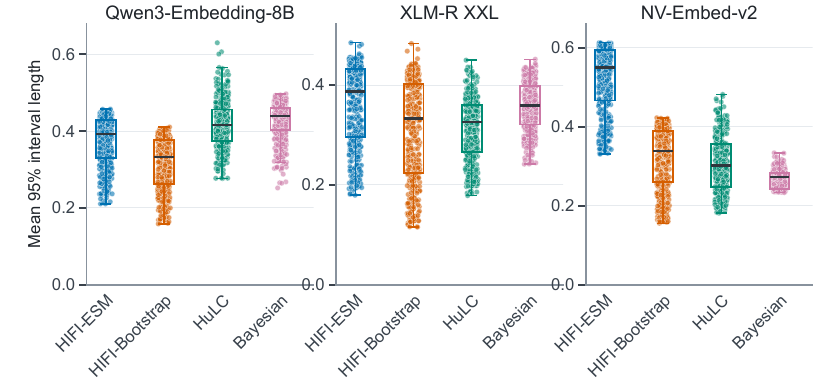}
\caption{
Target-wise mean lengths of nominal 95\% pointwise confidence
intervals for the text probability-prediction task. Each
point represents one evaluation target and reports the confidence-interval length averaged across 100 replications. Boxplots summarize the distribution of mean interval lengths across targets.
}
\label{fig:text-target-width}
\end{figure}

Figure~\ref{fig:text-target-coverage} reports empirical 95\%
coverage across individual evaluation targets, while Figure~\ref{fig:text-target-width} summarizes the corresponding
distribution of mean 95\% interval lengths. HIFI-ESM and
HIFI-Bootstrap exhibit relatively stable target-wise calibration
across all three language backbones, with coverage concentrated near the nominal level for most targets. HulC shows under-coverage for a subset of targets, whereas Bayesian-PEFT exhibits near-complete coverage under XLM-R XXL but pronounced under-coverage for many targets under Qwen3-Embedding-8B and especially NV-Embed-v2. In terms of interval length, HIFI-Bootstrap generally produces shorter intervals than HIFI-ESM. Notably, the shorter intervals produced by Bayesian-PEFT under NV-Embed-v2 coincide with substantial target-wise under-coverage. These results show that HIFI provides well-calibrated uncertainty quantification across evaluation targets.

\begin{figure}[t]
\centering
\includegraphics[width=\textwidth]
{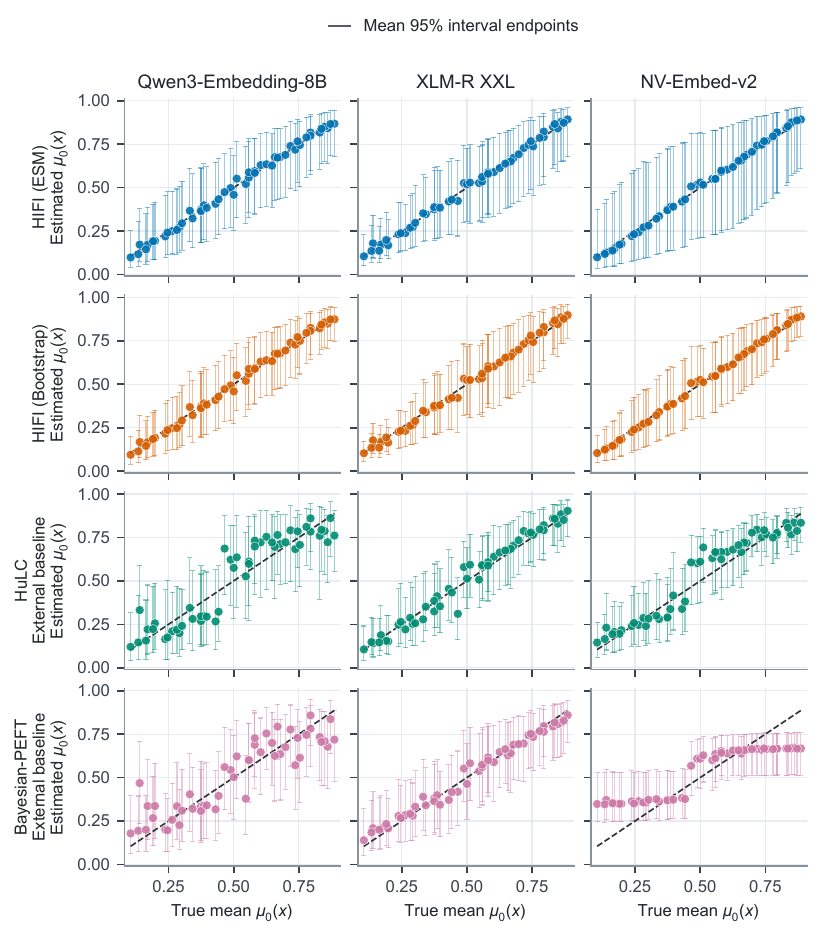}
\caption{
Target-level probability estimation and uncertainty for 50 evaluation
targets selected by stratified random sampling across languages and
true-probability bins. Columns correspond to the three language
backbones, and rows correspond to HIFI-ESM, HIFI-Bootstrap, HulC,
and Bayesian-PEFT. Points show the estimated conditional probability
averaged across replications for each target, and vertical
segments show the corresponding mean lower and upper endpoints of the
95\% intervals. The dashed diagonal line represents perfect
agreement with the true conditional probability $\mu_0(x)$.
}
\label{fig:text-target-predictions}
\end{figure}

For visualization, 50 targets are selected from the evaluation panel by stratified random sampling across languages and true-probability bins. Figure~\ref{fig:text-target-predictions} shows the corresponding mean point estimates and mean lower and upper endpoints of the 95\% intervals across replications. Across all three backbones, the HIFI estimators closely track the identity line over the range of true conditional probabilities. The external baselines show more pronounced
target-specific bias, with HulC deviating toward the center of the probability range for some targets and Bayesian-PEFT exhibiting particularly strong systematic deviations under NV-Embed-v2.

\subsection{Ablation Study 1}
\label{app:ablation-stage1-size}

\begin{figure}[t]
\centering
\includegraphics[width=\textwidth]{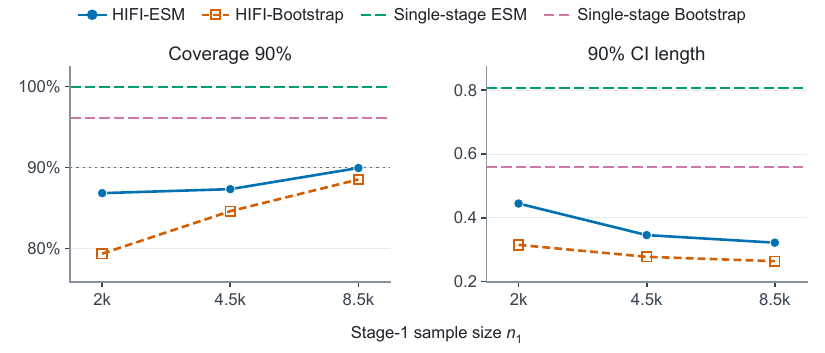}
\caption{
Effect of Stage~1 task-adaptation sample size on 90\% coverage and CI length. The dotted line marks the nominal 90\% coverage level.
}
\label{fig:ablation-s1-90}
\end{figure}

We provide complementary results on the effect of Stage~1
task-adaptation. Figure~\ref{fig:ablation-s1-90} shows the same qualitative pattern at the 90\% confidence level. Holding the Stage~2 inference sample fixed, increasing $n_1$ moves empirical coverage toward the nominal 90\% level while reducing mean interval
length for both ESM and bootstrap. The corresponding single-stage baselines attain conservative coverage, but with substantially larger intervals.

\begin{figure}[t]
\centering
\includegraphics[width=\textwidth]{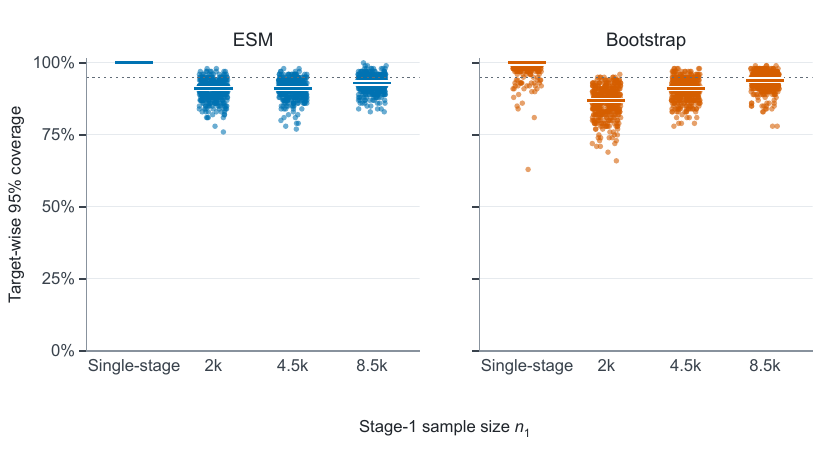}
\includegraphics[width=\textwidth]{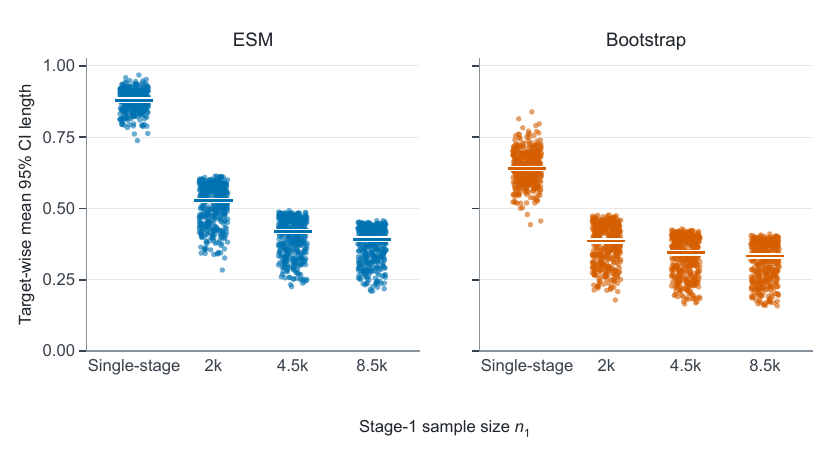}
\caption{
Target-wise 95\% coverage (top) and mean 95\% confidence-interval length (bottom) for ESM and bootstrap across the single-stage baseline and HIFI with $n_1\in\{2{,}000,4{,}500,8{,}500\}$. Each point
represents one of the 424 evaluation targets, with coverage and
interval length computed across 100 replications.
}
\label{fig:ablation-s1-target-distributions}
\end{figure}

Figure~\ref{fig:ablation-s1-target-distributions} shows similar target-wise patterns. For ESM and bootstrap, HIFI coverage is below 95\% at smaller
$n_1$ and moves closer to the nominal level at $n_1=8{,}500$. The interval lengths decrease across most targets as $n_1$ increases, while the single-stage baselines produce substantially wider intervals.

\begin{figure}[t]
\centering
\includegraphics[width=\linewidth]{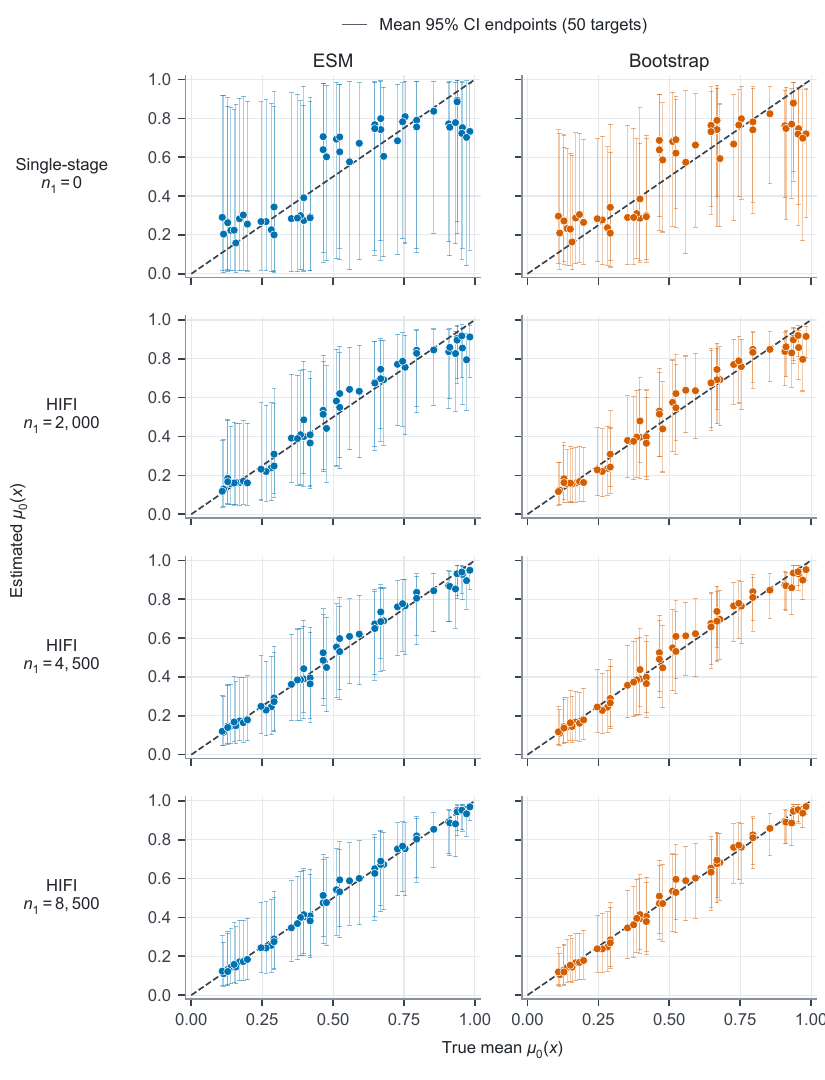}
\caption{
Target-wise point estimates and mean endpoints of 95\% confidence intervals for 50 representative targets, selected by stratified random sampling across languages and true-probability bins. Columns correspond to ESM and bootstrap; rows show the single-stage baseline and HIFI with $n_1\in\{2{,}000,4{,}500,8{,}500\}$. Points denote estimates averaged over 100 replications, and vertical segments connect the mean lower and upper confidence-interval endpoints. The dashed line is the identity line.
}
\label{fig:ablation-s1-target-estimates}
\end{figure}

Figure~\ref{fig:ablation-s1-target-estimates} further compares point
estimates and mean 95\% interval endpoints for 50 representative
targets. The single-stage fits show larger deviations from the identity
line and much wider intervals. Under HIFI, the point estimates track
the truth more closely as $n_1$ increases, while the corresponding
intervals become progressively narrower. These target-level results
are consistent with the aggregate findings in
Figure~\ref{fig:ablation-s1-main}.

\subsection{Ablation Study 2}
\label{app:ablation_direct}

\begin{figure}[t]
    \centering
    \includegraphics[width=\linewidth]{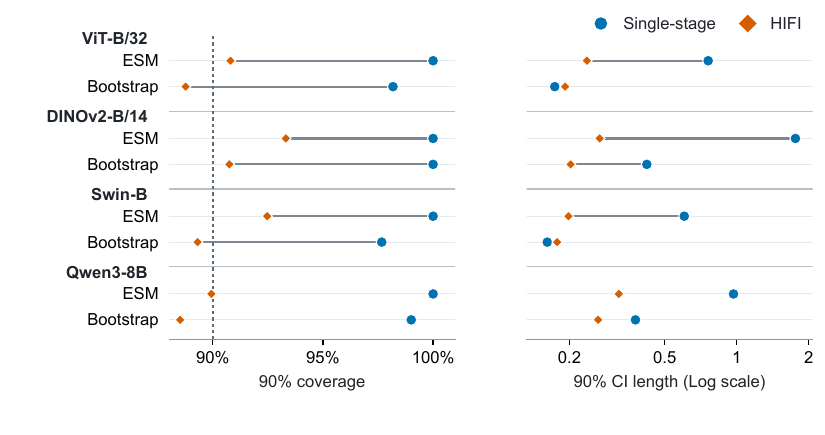}
    \caption{
    HIFI versus direct single-stage resampling under the same total
    data budget at the 90\% confidence level.
    }
    \label{fig:ablation-direct-90}
\end{figure}

Figure~\ref{fig:ablation-direct-90} reports the corresponding results at the 90\% confidence level. Consistent with the 95\% results in Figure~\ref{fig:ablation-direct}, direct single-stage resampling remains strongly conservative across backbones and methods, whereas HIFI produces coverage closer to the nominal level with substantially shorter intervals in most settings.

\begin{figure}[t]
    \centering
    \includegraphics[width=\linewidth]{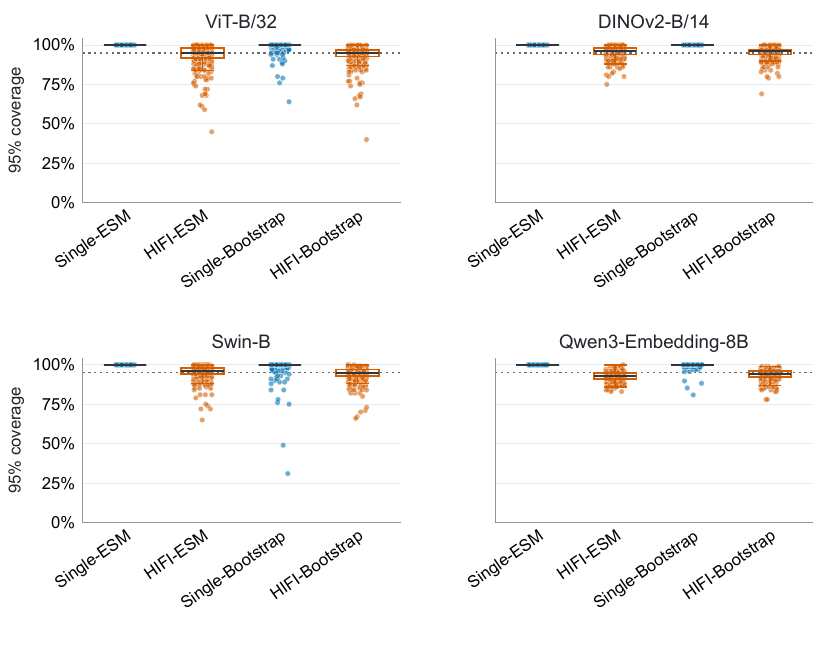}
    \caption{
    Target-wise empirical coverage of nominal 95\% confidence
    intervals for HIFI and direct single-stage resampling across
    backbones and inference methods. Each point represents one fixed
    evaluation target.
    }
    \label{fig:ablation-direct-target-coverage}
\end{figure}

\begin{figure}[t]
    \centering
    \includegraphics[width=\linewidth]{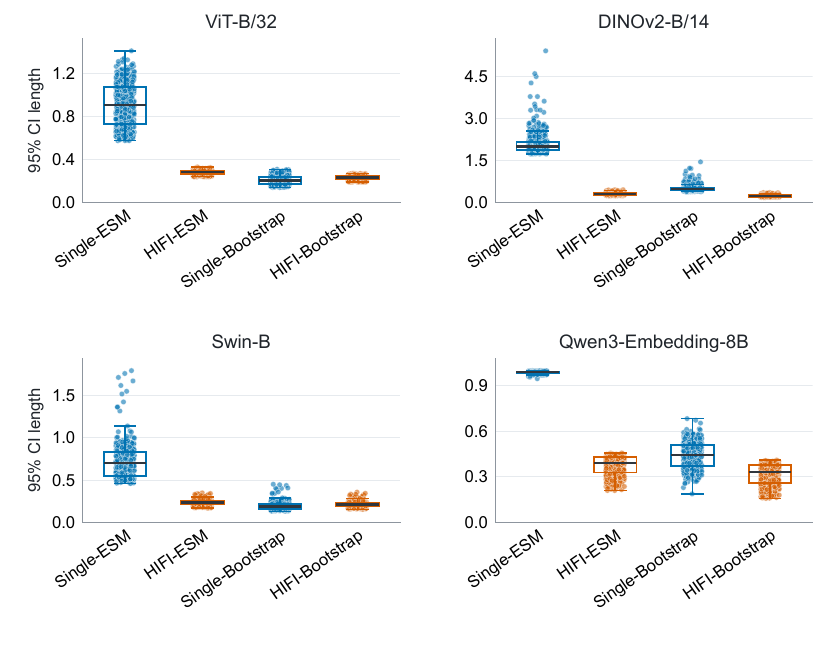}
    \caption{
    Target-wise lengths of nominal 95\% confidence intervals for
    HIFI and direct single-stage resampling across backbones and
    inference methods.
    }
    \label{fig:ablation-direct-target-width}
\end{figure}

Figures~\ref{fig:ablation-direct-target-coverage} and
\ref{fig:ablation-direct-target-width} provide target-level results at the 95\% confidence level. The single-stage procedures exhibit coverage concentrated near 100\% across most targets, while HIFI coverage is concentrated closer to the nominal 95\% level. HIFI produces substantially shorter intervals for ESM across all
four backbones. For bootstrap, HIFI yields shorter intervals for
DINOv2-B/14 and Qwen3-Embedding-8B, with comparable interval
lengths for ViT-B/32 and Swin-B.

\subsection{Computational Cost}
\label{app:computation}

Table~\ref{tab:training_compute_runtime} reports training EFLOPs and wall-clock time per replication for HIFI and the single-stage configurations. Vision experiments were run on A40 or RTX 3090 GPUs, while the language-model experiments used A100 80GB or B200 GPUs. All reported experiments use a single GPU.

Training EFLOPs are estimated analytically from the model architecture and executed workload, counting one multiply--add as two FLOPs. The calculation includes forward computation through the backbone, LoRA adapters, and prediction heads; backward computation for activation and trainable-parameter gradients; loss computation; and AdamW updates.
For HIFI, the total includes the Stage~1 training cost once together
with the cost of all Stage~2 ensemble or bootstrap members. For the
single-stage configurations, it includes training all ensemble or
bootstrap members directly on the responses using the combined
training sample. We report mean training EFLOPs per replication.

\begin{table}[t]
\centering
\caption{
Training compute and wall-clock time for HIFI and single-stage configurations across tasks, backbones, and inference methods.
}
\label{tab:training_compute_runtime}
\small
\setlength{\tabcolsep}{8pt}
\renewcommand{\arraystretch}{1.12}

\begin{tabular}{lllcrr}
\toprule
\textbf{Backbone}
& \textbf{Method}
& \textbf{Procedure}
& \textbf{GPU}
& \shortstack{\textbf{Training}\\\textbf{EFLOPs/rep}}
& \shortstack{\textbf{Wall-clock}\\\textbf{time/rep}} \\
\midrule

ViT-B/32
& ESM
& Single-stage
& A40
& 0.2934
& 2 h 44 min \\
& ESM
& HIFI
& RTX 3090
& 0.0239
& \textbf{15 min} \\
& Bootstrap
& Single-stage
& A40
& 0.4685
& 2 h 46 min \\
& Bootstrap
& HIFI
& RTX 3090
& 0.0668
& \textbf{32 min} \\

\addlinespace[4pt]

DINOv2-B/14
& ESM
& Single-stage
& A40
& 0.3174
& 1 h 48 min \\
& ESM
& HIFI
& RTX 3090
& 0.0649
& \textbf{37 min} \\
& Bootstrap
& Single-stage
& A40
& 1.0715
& 6 h 04 min \\
& Bootstrap
& HIFI
& RTX 3090
& 0.1720
& \textbf{1 h 01 min} \\

\addlinespace[4pt]

Swin-B
& ESM
& Single-stage
& A40
& 0.2239
& 1 h 03 min \\
& ESM
& HIFI
& RTX 3090
& 0.0312
& \textbf{24 min} \\
& Bootstrap
& Single-stage
& A40
& 0.4497
& 2 h 20 min \\
& Bootstrap
& HIFI
& RTX 3090
& 0.0604
& \textbf{34 min} \\

\addlinespace[4pt]

Qwen3-Embedding-8B
& ESM
& Single-stage
& A100 80GB
& 5.9344
& 7 h 37 min \\
& ESM
& HIFI
& A100 80GB
& 0.9394
& \textbf{1 h 21 min} \\
& Bootstrap
& Single-stage
& A100 80GB
& 10.1817
& 13 h 00 min \\
& Bootstrap
& HIFI
& A100 80GB
& 1.3349
& \textbf{1 h 51 min} \\

\addlinespace[4pt]

XLM-R XXL
& ESM
& HIFI
& A100 80GB
& 1.4369
& \textbf{2 h 00 min} \\
& Bootstrap
& HIFI
& A100 80GB
& 2.8516
& \textbf{3 h 47 min} \\

\addlinespace[4pt]

NV-Embed-v2
& ESM
& HIFI
& B200
& 1.8604
& \textbf{55 min} \\
& Bootstrap
& HIFI
& B200
& 6.4052
& \textbf{2 h 59 min} \\

\bottomrule
\end{tabular}

\vspace{3pt}
\begin{minipage}{0.97\textwidth}
\footnotesize
\textit{Note.}
All experiments were run on a single GPU.
Wall-clock time and EFLOPs are reported per replication.
\end{minipage}
\end{table}

\section{Additional Real-Data Analysis Details and Results}
\label{app:xenium}
We provide additional details on the Xenium data analysis.

\subsection{Data preprocessing and splitting}

We use two consecutive sections from the same breast cancer specimen
in the Xenium dataset of \citet{janesick2023high}, following the
cross-section prediction setting of iStar
\citep{zhang2024inferring}. Section~1 (GSM7780153) provides the
source data for model fitting, whereas Section~2 (GSM7780154) is
held out entirely for evaluation. We retain transcripts with quality
scores of at least 20 and use all 313 genes.

Following the iStar preprocessing pipeline, the registered H\&E
images are processed at $0.5\,\mu\mathrm{m}$ per image pixel.
The native prediction grid consists of $8\times8\,\mu\mathrm{m}$
superpixels. Supervision is provided through pseudo-Visium spots
with diameter $55\,\mu\mathrm{m}$ and center-to-center spacing
$100\,\mu\mathrm{m}$, each associated with 37 native superpixels.

The 4,455 source-section spots are partitioned into spatial blocks,
and the blocks are randomly assigned to the Stage~1 and Stage~2
training and validation sets. Predictions are produced natively at
$8\,\mu\mathrm{m}$ resolution and aggregated to $16\,\mu\mathrm{m}$
blocks for reporting. Training and resampling operate at the
pseudo-Visium-spot level. Table~\ref{tab:xenium-settings} summarizes
the principal settings.

\subsection{Model implementation and inference}

\paragraph{Model implementation} HIFI uses the pretrained hierarchical vision transformer backbone of
iStar. Each superpixel combines pretrained HViT features with mean RGB
intensities and spatial coordinates. RGB intensities are divided by
255, and coordinates are scaled to $[-1,1]$ using each section's
tissue bounding box.

Following iStar, the model is trained through weak supervision at
the pseudo-Visium-spot level. For each gene $g$, let $a_g$ and $s_g$ denote the minimum and range,
respectively, of its expression counts across the Stage~1 training spots. These quantities are used to normalize the targets in
both stages and are kept fixed throughout model fitting and inference.
Let $\widetilde Y_s\in\mathbb R^{313}$ denote the resulting normalized
expression vector for spot $s$.

Let $\mathcal M_s$ denote the set of 37 native superpixels associated
with spot $s$, and let $f_\theta(X_u)\in\mathbb R^{313}$ denote the
normalized gene-expression prediction for superpixel $u$. The
spot-level prediction is
$\bar f_\theta(s)=|\mathcal M_s|^{-1}\sum_{u\in\mathcal M_s}f_\theta(X_u)$,
and model fitting minimizes
$\mathcal L(\theta)=(313|\mathcal S|)^{-1}
\sum_{s\in\mathcal S}\|\widetilde Y_s-\bar f_\theta(s)\|_2^2$.
Thus, spot-level measurements provide weak supervision while the model
retains predictions on the finer superpixel grid.

We apply HIFI using the two-stage fitting and inference procedure, with model fitting and resampling performed at the pseudo-Visium-spot level. Model architecture, PEFT, optimization, and other implementation settings are summarized in Table~\ref{tab:xenium-settings}.

\begin{table}[tbp]
    \centering
    \caption{HIFI settings for the Xenium experiment.
    Batch sizes and resampling sample sizes count pseudo-Visium spots.}
    \label{tab:xenium-settings}
    \small
    \renewcommand{\arraystretch}{1.10}
    \begin{tabularx}{\linewidth}{@{}
        >{\raggedright\arraybackslash}p{0.25\linewidth}
        >{\raggedright\arraybackslash}X
        >{\raggedright\arraybackslash}X@{}}
        \toprule
        Shared setting & \multicolumn{2}{l}{Specification} \\
        \midrule
        Input features &
            \multicolumn{2}{l}{
                192 HViT-4K + 384 HViT-256 + 3 RGB + 2 coordinates} \\
        MLP layer dimensions &
            \multicolumn{2}{l}{
                $581\to256\to256\to256\to256\to313$} \\
        Hidden activation &
            \multicolumn{2}{l}{
                Leaky ReLU with negative slope 0.1} \\
        Trainable MLP layers &
            \multicolumn{2}{l}{
                All hidden layers and the output layer} \\
        LoRA targets &
            \multicolumn{2}{l}{
                 QKV and attention output projections} \\
        LoRA parameters &
            \multicolumn{2}{l}{
                Rank 8; scaling 16; dropout 0} \\
        Optimizer &
            \multicolumn{2}{l}{
                Adam; learning rate $10^{-4}$; weight decay 0} \\
        Training batch size &
            \multicolumn{2}{l}{8} \\
        \midrule
        Stage-specific setting & Stage~1 & Stage~2 \\
        \midrule
        Training/validation spots &
            2,495 / 624 &
            1,069 / 267 \\
        HViT-256 adaptation &
            LoRA in all 12 blocks &
            Additional member-specific LoRA in the last 3 blocks \\
        MLP initialization &
            Random initialization &
            Initialized from Stage~1 hidden-layer weights;
            final layer initialized to zero \\
        MLP output activation &
            $\operatorname{ELU}_{0.01}(t)+0.01$ &
            Identity \\
        \midrule
        \multicolumn{3}{@{}l}{\textit{Stage~2 resampling}} \\
        ESM &
            \multicolumn{2}{l}{
                500 subsets of 532 out of 1,069 spots;
                without replacement} \\
        Bootstrap &
            \multicolumn{2}{l}{
                500 samples of 1,069 spots;
                with replacement} \\
        \bottomrule
    \end{tabularx}
\end{table}

\paragraph{Statistical inference} Let $p^{(b)}_{ug}$ denote the final normalized prediction of member $b$
for gene $g$ at native superpixel $u$. Using the Stage~1 normalization
parameters defined above, we convert this prediction to a native count
contribution as
$c^{(b)}_{ug}=(s_g p^{(b)}_{ug}+a_g)/37$, where the division by 37
accounts for the 37 superpixels contributing to each pseudo-Visium spot.

For each nonoverlapping $16\,\mu\mathrm{m}$ block $j$, let
$\mathcal A_j$ contain its valid native superpixels.  We first aggregate within each member and then
compute the ensemble point estimate:
\begin{equation}
    T^{(b)}_{jg}
    = \sum_{u\in\mathcal A_j} c^{(b)}_{ug},
    \qquad
    \widehat\mu_{jg}
    = \frac{1}{B}\sum_{b=1}^{B}T^{(b)}_{jg},
    \qquad B=500.
    \label{eq:xenium-member-aggregation}
\end{equation}
These operations are performed separately for ESM and bootstrap.
For ESM, we apply the ESM variance estimator to the aggregated
member predictions and their source-spot inclusion indicators.
Writing the resulting variance estimate as
$\widehat V^{\mathrm{ESM}}_{jg}$, the nominal 95\% interval is
\begin{equation}
    \left[
    \widehat\mu_{jg}
        -z_{0.975}\sqrt{\widehat V^{\mathrm{ESM}}_{jg}},\;
    \widehat\mu_{jg}
        +z_{0.975}\sqrt{\widehat V^{\mathrm{ESM}}_{jg}}
    \right].
    \label{eq:xenium-esm-interval}
\end{equation}
For bootstrap, the interval endpoints are the empirical 2.5th
and 97.5th percentiles of $\{T^{(b)}_{jg}\}_{b=1}^{B}$.

\subsection{Evaluation}

We rerun the official iStar implementation on all 4,455
source-section spots using its released image features. Each run
uses the official median aggregation of five model states, each
trained for 400 epochs. We perform five runs with different
optimization randomness. The main figure~\ref{fig:xenium-main} displays the spatial
prediction from the first run.

All methods are evaluated on the same set of 107,052 registered
$16\,\mu\mathrm{m}$ blocks containing at least one valid native
superpixel. Observed and predicted expression are evaluated as block-total counts.

We evaluate point-prediction performance using root mean squared
error (RMSE), Pearson correlation coefficient (PCC), and structural
similarity index (SSIM). RMSE and PCC are computed directly from
the observed and predicted block-total count maps. SSIM follows
the released iStar implementation, using uniform $7\times7$
windows, sample covariance, and only windows fully contained in
the registered support. For each gene, the SSIM dynamic range is
defined by the observed count range and held fixed across all
methods and iStar runs.

For each gene and metric, the iStar reference is the average over
the five runs, $M_g^{\mathrm{iStar}}
= 5^{-1}\sum_{r=1}^{5} M_g^{(r)}$. The scatter plots contain one
point per gene and compare these reference values with the
corresponding HIFI scores.

\subsection{Additional results and interpretation}

\begin{figure}[t]
    \centering
    \includegraphics[width=\linewidth]{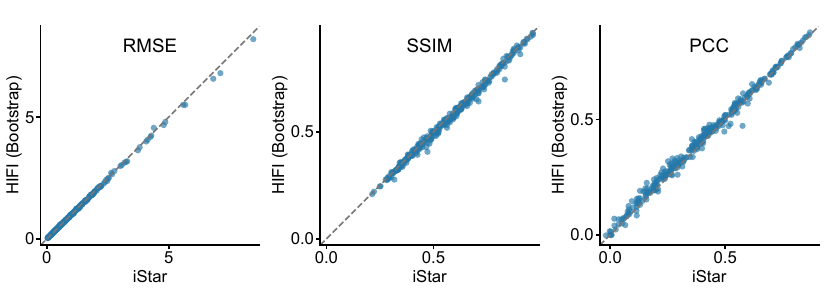}
    \caption{
    Point-prediction comparison between HIFI with bootstrap and
    iStar at $16\,\mu\mathrm{m}$ resolution. Each point represents
    one of the 313 genes, and dashed lines indicate equality.
    Lower RMSE and higher SSIM and PCC indicate better agreement
    with observed expression. 
    }
    \label{fig:xenium-bootstrap-comparison}
\end{figure}

Figure~\ref{fig:xenium-bootstrap-comparison} compares bootstrap
with iStar across all 313 genes and shows comparable
point-prediction performance.

\begin{figure}[t]
    \centering
    \includegraphics[width=\linewidth]{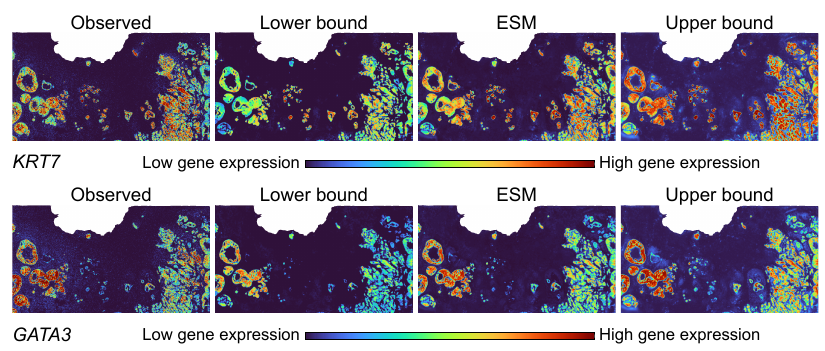}
    \caption{
    Additional ESM interval maps for \textit{KRT7} and
    \textit{GATA3}. Columns show observed expression, the lower
    endpoint, the point estimate, and the upper endpoint of
    the nominal 95\% pointwise interval for mean expression.
    The corresponding \textit{TACSTD2} example appears in
    Figure~\ref{fig:xenium-main}.
    }
    \label{fig:xenium-esm-intervals}
\end{figure}

\begin{figure}[t]
    \centering
    \includegraphics[width=\linewidth]{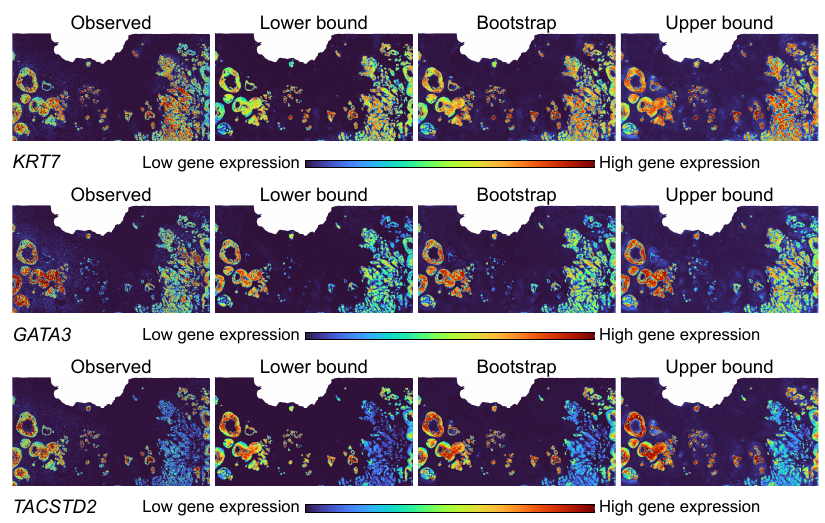}
    \caption{
    Bootstrap interval maps for \textit{KRT7}, \textit{GATA3},
    and \textit{TACSTD2}. Columns show observed expression,
    the lower percentile endpoint, the bootstrap mean, and
    the upper percentile endpoint. The endpoints correspond
    to nominal 95\% pointwise intervals. Each gene uses the
    same display mapping as its ESM panels.
    }
    \label{fig:xenium-bootstrap-intervals}
\end{figure}

Figure~\ref{fig:xenium-esm-intervals} presents additional genes
with their ESM confidence interval maps, while
Figure~\ref{fig:xenium-bootstrap-intervals} provides three additional
visual examples of bootstrap confidence interval maps. These spatial maps illustrate the uncertainty quantification
provided by HIFI on a real scientific prediction task.

\end{document}

%% file: math_commands.tex
\usepackage{amsmath,amsfonts,bm}

\def\eqref#1{equation~\ref{#1}}
\def\1{\bm{1}}

\DeclareMathAlphabet{\mathsfit}{\encodingdefault}{\sfdefault}{m}{sl}
\SetMathAlphabet{\mathsfit}{bold}{\encodingdefault}{\sfdefault}{bx}{n}